%% file: main.tex
\documentclass[10pt,twocolumn,letterpaper]{article}

\usepackage{iccv}              
\usepackage{url}
\usepackage{graphicx}
\usepackage{booktabs}
\usepackage{array}
\usepackage{amsmath}  
\usepackage{caption}
\usepackage{tikz}
\usepackage{mdframed}
\usepackage{booktabs}
\usepackage[normalem]{ulem}
\usepackage{enumitem}
\usepackage[accsupp]{axessibility}
\useunder{\uline}{\ul}{}

\makeatletter
\renewcommand{\paragraph}{%
 \@startsection{paragraph}{4}{0pt} 
 {1.7ex plus 0.8ex minus 0.2ex}  
 {-1em}  
 {\normalfont\normalsize\bfseries} 
}
\makeatother


\definecolor{cvprblue}{rgb}{0.21,0.49,0.74}
\usepackage[pagebackref,breaklinks,colorlinks,allcolors=cvprblue]{hyperref}
\usepackage{placeins}

\usepackage{marvosym}

\def\paperID{9310} 
\def\confName{ICCV}
\def\confYear{2025}

\title{Bitrate-Controlled Diffusion for Disentangling Motion and Content in Video}

\author{
Xiao Li$^{1\dagger}$\quad 
Qi Chen$^{1,2,3\dagger}$\quad 
Xiulian Peng$^{1}$\quad
Kai Yu$^{2}$\quad 
Xie Chen$^{2,3}$\textsuperscript{\Letter}\quad
Yan Lu$^{1}$\textsuperscript{\Letter}\\
$^1$Microsoft Research Asia\quad $^2$Shanghai Jiao Tong University\quad $^3$Shanghai Innovation Institute\\
{\tt\small \{cq1073554383, kai.yu, chenxie95\}@sjtu.edu.cn\quad \{xili11, xipe, yanlu\}@microsoft.com\quad}
}

\begin{document}
\maketitle
\input{sec/0_abstract}    
\footnotetext{$\dagger$ Joint first authors. The major part of the work was done when Qi Chen was an intern at MSRA.}
\footnotetext{\textsuperscript{\Letter} Corresponding authors.}
\input{sec/1_intro_arxiv}
\input{sec/2_related}
\input{sec/3_method}
\input{sec/4_experiments}
\input{sec/5_conclusion}
\input{sec/7_acknowledgement}
{
    \small
    \bibliographystyle{ieeenat_fullname}
    \bibliography{main}
}
\renewcommand{\thefigure}{A\arabic{figure}}
\renewcommand{\thetable}{A\arabic{table}}

\clearpage
\input{sec/6_suppl}


\end{document}

%% file: sec/0_abstract.tex
\begin{abstract}
We propose a novel and general framework to disentangle video data into its dynamic motion and static content components. Our proposed method is a self-supervised pipeline with less assumptions and inductive biases than previous works: it utilizes a transformer-based architecture to jointly generate flexible implicit features for frame-wise motion and clip-wise content, and incorporates a low-bitrate vector quantization as an information bottleneck to promote disentanglement and form a meaningful discrete motion space. 
The bitrate-controlled latent motion and content are used as conditional inputs to a denoising diffusion model to facilitate self-supervised representation learning.
We validate our disentangled representation learning framework on real-world talking head videos with motion transfer and auto-regressive motion generation tasks.
Furthermore, we also show that our method can generalize to other types of video data, such as pixel sprites of 2D cartoon characters.
Our work presents a new perspective on self-supervised learning of disentangled video representations, contributing to the broader field of video analysis and generation.
\end{abstract}
                                                                                                             

%% file: sec/1_intro_arxiv.tex
\section{Introduction}
\label{sec:intro}
Video data contains rich temporal information and intricate patterns of movement and change, offering a deeper understanding of real-world environments.
Following the intrinsic information structure, a compact and robust representation of video sequences is the disentangled representation in terms of the static part (i.e. content) and the dynamic part (i.e. motion).
The content part is highly semantic, encapsulating the unchanging aspects of the scene.
The motion component captures the time-varying essentials of the input data, embodying the essence of change and movement within the video. 
Hence, the disentangled representation offers a structured space for 
accurately modeling, processing, and understanding the motion and the content, enabling the interpretation, the synthesis and the manipulation of the visual world in its first principles.
\input{figures/overview}

Despite the clear advantages of the disentangled representation of video data, the disentanglement learning task itself remains challenging. The decomposition task is intrinsically an underdetermined problem, not to mention the extremely high-dimensional nature of video signals which exacerbates the computational demands as well as introduces complexities in the construction of meaningful disentanglement.
Existing works either introduce additional assumptions~\cite{li2018disentangled, bai2021contrastively, naiman2023sample, berman2024sequential, simon2024sequential, mallya2022implicit} or impose task-specific prior constraints~\cite{wang2021one,siarohin2019first, li2021motion,gao2023high, zhu2020s3vae, guo2024liveportrait, liu2024anitalker}.
These additional assumptions are often overly idealistic and limit the expressiveness of disentanglement features in real scenarios, while the introduced priors constrain the target framework to specific tasks, restricting its applicability to a narrow domain.

In this paper, we propose a general and practical framework for disentangling motion and content representations in real-world video data, minimizing assumptions and task-specific inductive biases.
The form of our disentangled representation is uniquely flexible - a purely implicit frame-wise motion feature and clip-wise global content latent features for a given video input sequence.
Compared with other explicit or hybrid representations, implicit features have less inductive bias, significantly increasing the expressiveness of our representation.
To make our disentanglement assumptions more general, we resort to the first principle and introduce an information bottleneck to regulate the information flow of implicit features during training. Inspired by recent neural codec methods~\cite{li2021deep}, we achieve this by leveraging a low-bitrate codebook.
Our key insight is that low-bitrate constraint acts as a general prior to enforce meaningful disentanglement, as video reconstruction under an information bottleneck shares the same goal as disentangled representations: providing a compact yet informative representation of the data~\cite{tishby2000information}. 
We facilitate the self-supervised video reconstruction under low-bitrate with a latent denoising diffusion model.
The usage of the latent diffusion model as a generative decoder for representation learning enhances the fidelity of the learned representation under information bottlenecks~\cite{hudson2023soda}.

Putting it all together, we propose \textbf{B}itrate \textbf{C}ontrolled \textbf{D}iffusion model, \textbf{BCD} for short, a new pipeline for learning disentangled motion and content representations from video sequences (\Cref{fig:overview}).
We demonstrate the expressiveness and the effectiveness of our proposed BCD framework by showing results on motion transfer and auto-regressive motion generation on a real world talking head video dataset~\cite{afouras2018lrs3}. 

Furthermore, we demonstrate the generality of our method by showing additional disentanglement results on other video data such as the LPC Sprites~\cite{reed2015deep}.
Our contributions can be summarized as follows:
\begin{itemize}[topsep=0pt]
    \item A novel self-supervised framework for disentangling motion and content in video with minimal assumptions and greater flexibility than prior methods.
    \item A low-bitrate vector quantization approach that serves as an information bottleneck to enforce meaningful disentanglement for real world videos.
    \item A structured and expressive latent video space enabled by a denoising diffusion model, supporting generative tasks such as motion transfer and auto-regressive video generation.
\end{itemize}

\paragraph{Ethics Statement.}
BCD is a research project that focuses exclusively on the self-supervised disentanglement of motion and content from video data. 
Our work strictly follows Microsoft’s AI principles. While we present several examples involving human faces, their sole purpose is to illustrate the effectiveness of disentangling real-world video representations in an intuitive and interpretable way. 
All human-related datasets used in this work are publicly available benchmarks that are widely adopted across academia and industry for evaluating facial motion transfer. Our use of these datasets follows established community practice to ensure comparability and alignment with prior work.
Our proposed framework is general-purpose: we did not develop or employ any human-specific technologies for deepfake generation, nor do we intend to generate content for misleading or deceptive purposes. At present, we have no plans to integrate this technology into any product or to make its implementation publicly available.

%% file: figures/overview.tex
\begin{figure*}[tb]
    \centering
    \includegraphics[width=\textwidth]{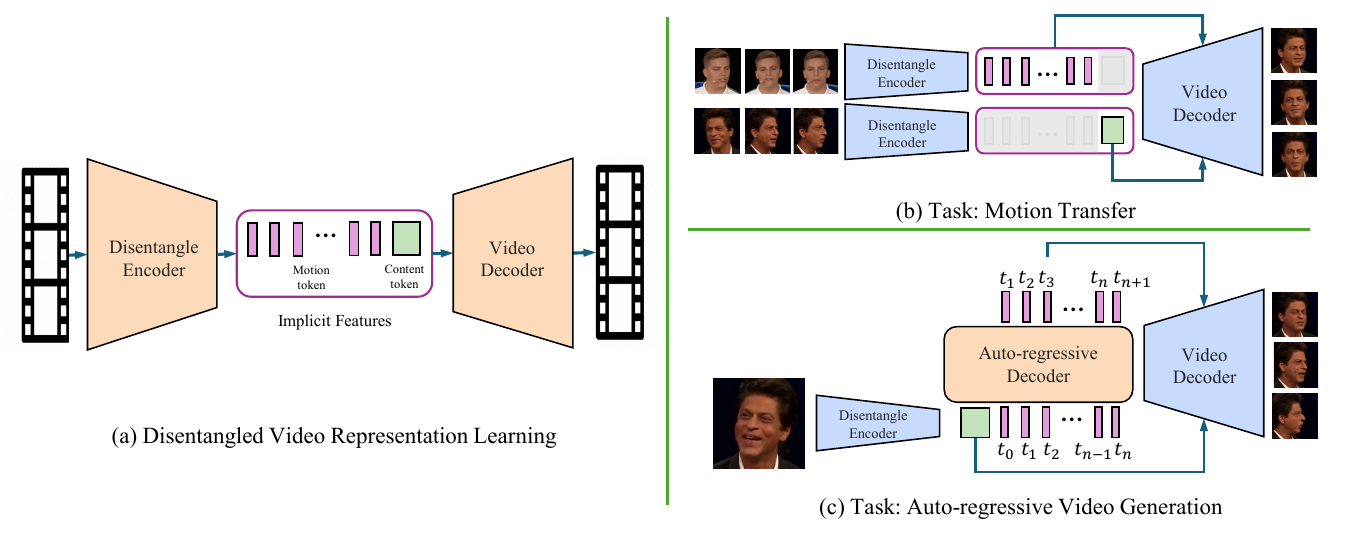}
    \caption{Overview of our method. (a) BCD learns compact and disentangled features to represent video data in a self-supervised manner: each video clip is represented by a clip-wise content feature and frame-wise motion feature sequences.
    (b) The learned disentangled video space natively supports editing tasks, e.g., swapping the content features between two videos for motion transfer.
    (c) The learned representations form a structured motion space, which further enables training of auto-regressive video generation.}
    \label{fig:overview}
\end{figure*}

%% file: sec/2_related.tex
\section{Related Works}
\label{sec:related_works}
\paragraph{\textbf{Visual Representation Learning}}
Recent years have witnessed significant advancements in image representation learning with self-supervised pre-training. 
The Vision Transformer (ViT)~\cite{dosovitskiy2020image} paired with pretask of contrastive learning~\cite{he2020momentum,chen2020simple,grill2020bootstrap} or masked image modeling~\cite{xie2022simmim,peng2022beit} encodes image patches into discrete tokens and leveraging Transformer architectures~\cite{vaswani2017attention} typically used in NLP for visual tasks.
Concurrently, generative techniques like VQ-VAE~\cite{van2017neural}, GANs~\cite{goodfellow2014generative}, VQ-based diffusion~\cite{gu2022vector,esser2021taming} or auto-regressive models~\cite{wu2022nuwa,yan2021videogpt} have demonstrated remarkable capabilities in image synthesis and editing. 
Specifically, recent studies~\cite{hudson2023soda, li2023self} have demonstrated the potential of diffusion models in learning robust visual representations. 
The emergence of video diffusion models such as \textit{Sora}~\cite{videoworldsimulators2024} also significantly advances the progress in video representation.
These techniques learned to compress high-dimensional visual data into lower-dimensional latent spaces while preserving essential semantic information.
Our approach leverages these pioneer works, aims at representing video in a \textit{disentangled} manner.

\paragraph{\textbf{Disentangled Video Representation}}
There are many works that attempts to separate motion and content from video sequences.
One line of works studies the disentanglement problem from a general perspective. They introduce additional assumptions within the VAE framework such as low-dimensional features for dynamics~\cite{li2018disentangled}, explicit dependence~\cite{simon2024sequential} or independence~\cite{bai2021contrastively, naiman2023sample, han2021disentangled} between motion and content, or that static content can be fully modeled from a single frame in the video~\cite{berman2024sequential}.
SKD~\cite{berman2023multifactor} explores multi-factor disentanglement with structured Koopman autoencoders, yet they only demonstrate preliminary results on toy datasets.
Our method avoids predefining motion-content correlations, allowing the aggregation of content from multiple frames, and regulate with bitrate control rather than explictly using feature with lower channels. These design choices result in a simple yet effective framework capable of handling real-world videos with complex motion and content.

One crucial real application for video disentanglement is the generation and editing of talking head videos. 
Existing methods often employing explicit optical flow for motion modeling~\cite{huang2022neural,zhu2020s3vae}, using specific priors such as landmarks, keypoints or 3D reconstructions~\cite{siarohin2019first,wang2021one,mallya2022implicit,hong23implicit, li2021motion,gao2023high,bounareli2023hyperreenact, grassal2022neural, mallya2022implicit, guo2024liveportrait}. LIA~\cite{wang2021latent} proposes to model image animation in a latent space using linear motion decomposition with a learnable orthogonal motion basis. 
Our goal is to explore a more general paradigm for motion-content disentanglement with minimal task-specific inductive bias.
While we include talking head synthesis as an evaluation task, our approach is not tailored to solve the talking head task exclusively. 

\paragraph{\textbf{Information Bottleneck and Neural Codec}}
The theoretical feasibility of our method is based on the information bottleneck (IB) theory.
The information bottleneck (IB) provides a principled framework for learning simplified yet informative representations of data~\cite{tishby2000information}. The core of IB is to retain the most relevant information of the data while discarding irrelevant details through a compression process. As noted in~\cite{alemi2016deep,achille2018emergence}, the use of IB not only yields a compact representation but also facilitates disentanglement.
Our approach leverages this intrinsic property of IB as a powerful tool to disentangle video into motion and semantic content. 

State-of-the-art neural video codecs~\cite{li2021deep,sheng2022temporal,li2022hybrid,li2023neural} have demonstrated promising results in efficient of video signals. These methods employ a strong IB with extremely high compression ratios, often relying on explicit motion vectors, such as optical flow. However, they generally do not explicitly model the disentanglement of content and motion.
Recently, disentangled low-bitrate audio codecs~\cite{jiang2023disentangled} have demonstrated that a low-bitrate vector quantizer can serve as a strong information bottleneck (IB) for separating speech content from speaker identity. Inspired by these advancements in neural codecs, we apply low-bitrate constraints as an IB for video disentanglement, but with a different objective—discovering disentangled representations of motion and content within an implicit latent space.

%% file: sec/3_method.tex
\section{Method}
\label{sec:method}
BCD is a self-supervised diffusion model that learns to encode video sequences into a disentangled representation of a frame-wise motion feature sequence and a sequence-wise content feature.
The overall framework of BCD is illustrated in \Cref{fig:pipeline}.
Following common latent diffusion approach~\cite{rombach2022high}, an input video sequence are first tokenized into a pre-frame latent sequence $z=\{z_t\: | \: t \in [1,T]\}$ with a pre-trained image VAE. 
Our BCD consists of a transformer encoder $\mathcal{T}(l)=(\{m_t\: | \: t \in [1,T]\}, c)$ that encodes the latent sequence $z$ into its disentangled representations of per-frame motion features $m=\{m_t\: | \: t \in [1,T]\}$ and a global content feature $c$ (\cref{sec:method:feature_extraction}).
To promote a correct disentanglement, 
the per-frame motion $m$ is bounded with a low-bitrate vector quantization bottleneck to retain only the essential motion information necessary for preserving the scene dynamics (\cref{sec:method:bitrate_control}).
These disentangled representations then serve as condition inputs to guide the reconstruction of the original latent $z$ through a diffusion model $\mathcal{D}$ (\cref{sec:method:diffusion}).
The full model is end-to-end trained with a rate-distortion objective (\cref{sec:method:training}). 
\input{figures/pipeline}

\subsection{Content and Motion Extraction}
\label{sec:method:feature_extraction}
The transformer architecture \cite{vaswani2017attention} is pivotal to our model for the extraction and assembly of information, owing to its proven capability to facilitate information transfer across varied modalities and domains with less inductive biases~\cite{li2023blip, mallya2022implicit}. 
In our framework, we utilize a transformer encoder $\mathcal{T}$ to simultaneously aggregate information among frames and build correlations between frames in the latent space.
To aggregate the content information, we prepend a fixed number of learnable queries with $K$ tokens, i.e., $q \in \mathbb{R}^{K \times d}$, to the feature sequence $\{z_t\: | \: t \in [1,T]\}$ as a prefix before input into the transformer network $\mathcal{T}$.
The learnable queries $q$ are optimized over the whole training set, so that it implicitly learns to robustly aggregate information from multiple video frames.
The output of the network is a prefixed latent sequence where the first $K$ prefix token corresponding to the content feature $c \in \mathbb{R}^{K \times C_c}$, followed by the motion sequence $m \in \mathbb{R}^{T \times C_m}$.
Compared to existing work which either uses image-based features with a fixed number, arbitrarily chosen reference frames, or simple pooling operators~\cite{gao2023high,mallya2022implicit,siarohin2019first},
our design can better encode content information of videos with large variations among frames (e.g., videos with different views or videos with certain details only exist in specific frames), and is flexible to the number of input frames.
We implement $\mathcal{T}$ with the T5~\cite{2020t5} transformer equipped with relative positional encoding.

\subsection{Feature Disentanglement}
\label{sec:method:bitrate_control}
Separating video latents into a global content feature and a frame-wise motion sequence using only the prefixed transformer approach in \cref{sec:method:feature_extraction}, does not inherently guarantee disentangled information encoding in the corresponding features.
Without constraints, the implicit features we use exhibit extremely high expressiveness, allowing arbitrary encoding of video information.
This leads to two common failure modes in disentanglement~\cite{berman2024sequential, simon2024sequential}: information preference, where the mutual information between inputs and latents is insufficient, and information leakage, where motion and content features become improperly entangled.

To ensure the disentanglement of the motion branch and content branch, we utilize Information Bottlenecks (IB) as a key component.
Our key observation is that both information leakage and information preference essentially stem from the improper distribution of information within the representation, failing to follow the correct disentanglement scheme.
According to the information bottleneck theory, modeling a representation through an IB shares the same ultimate goal to disentanglement representations, that is, to provide a most compact yet informative representation of the data~\cite{tishby2000information}.
On one hand, by seeking a compact representation, the IB forces the model to discard unnecessary details, which often corresponds to the entangled factors in the data~\cite{alemi2016deep}.
On the other hand, by preserving the most relevant information for a specific task, the model is encouraged to keep the factors that are truly meaningful and universal for the data generation process~\cite{achille2018emergence}.
Therefore, a robust information bottleneck can not only find a compact representation, but also guide the disentanglement process by aiding in separating the underlying factors of the data.
Inspired by recent work on low-bitrate neural codec~\cite{jiang2023disentangled}, we implement our IB by passing the per-frame motion features $\{m_t\}$ through a bitrate-controlled vector quantization (VQ) module, detailed as follows.

\paragraph{\textbf{Group VQ}} We adopt group VQ~\cite{baevski2019vq} to maximize the capacity upper bound of our codebooks. Specifically, the motion latent for input frame at time $t$, $m_t \in R^c$ is splited into $N$ groups $\{m_t^i \in R^{c/N} \: | \: i \in [1,N]\}$. Each group is then individually processed by a codebook $E^i$ with $K$ code entries of $c_{vq}=c/N$ channels. 
We use a distance-gumbel-softmax layer to quantize $m^i_t$ and sample from the codebook $E^i$.
The sampling distribution is based on the distance between input and codes: 
\begin{align}
d_t^i &= [\ell(m_t^i, e^i_1), \ell(m_t^i, e^i_2), \cdots, \ell(m_t^i, e^i_K)] \\
\mu_t^i &= GumbelSoftmax(-\alpha \cdot d^i_t)
\end{align}
where $\ell(\cdot, \cdot)$ measures the L2 distance, $\alpha$ is a scale factor, $e^i_j$ is the j-th entry of the i-th codebook group, and $\mu_t^i$ is the sampling distribution from $E^i$ for input $m^i_t$. Quantized code $\hat{m}^i_t$ are sampled according to $\mu_t^i$ with the gumbel reparametrization trick during training and retrieved with argmax of the distribution for inference.

\paragraph{Bitrate Control}
According to Shannon's source coding theorem~\cite{cover1999elements}, the metric for compressibility (or compactness) of the quantized latent motion feature $\hat{m}^i_t$ is the entropy:
\begin{equation}
\label{eq:shannon_entropy}
    \mathcal{H}(\hat{m}^i_t|E^i)=-\sum_{j}P(\hat{m}^i_t=e^i_j)log(P(\hat{m}^i_t=e^i_j))
\end{equation}
For training, since we have the sampling distribution $\mu_t^i$ for each input, we can approximate \cref{eq:shannon_entropy} with the average sampling histogram over the training batch $\mathcal{B}$, i.e.,
\begin{align}
    \mu_{avg}^i&=\frac{1}{|\mathcal{B}|}\sum_{\mathcal{B}}\mu_t^i\ , \  \forall \mu_t^i\in \mathcal{B} \\ 
    \hat{\mathcal{H}}(\hat{m}^i_t|E^i)&=-\sum\mu^i_{avg}log(\mu^i_{avg})
\end{align}
The motion bitrate per frame from the model can be computed as $\mathcal{H}_{model}=\sum_{i} \mathcal{H}(\hat{m}^i_t|E^i)$. 
To constrain the target bitrate, we employ MSE loss between $\mathcal{H}_{model}$ and a target bitrate $\mathcal{H}_{target}$. The target bitrate $\mathcal{H}_{target}$ is a hyper-parameter and we empirically set it slightly lower than the average bitrate of the video dataset (approximated from existing video codec methods).

\paragraph{Remarks}
The bitrate-controlled VQ enforces an optimal codebook that meets the average bitrate requirement for video sets while allowing the bitrate of individual videos to remain flexible, accommodating diverse inputs.
During training, the bitrate constraint is equipped with the data fidelity loss, forming a rate-distortion objective. 
According to rate-distortion theory, optimal compression inherently preserves expressiveness by minimizing distortion at a given bitrate.
By selecting an appropriately low but non-zero bitrate, our low-bitrate controller for motion helps alleviate both information leakage and information preference. Restricting the motion bitrate prevents content from leaking into motion, otherwise the leaked content information would cause the motion bitrate to exceed its constraint. Meanwhile, maintaining a nonzero target bitrate mitigates information preference by preventing excessive information loss in motion feature and ensuring reconstruction fidelity.

\subsection{Conditional Diffusion Decoder}
\label{sec:method:diffusion}
The BCD decoder is a conditional latent denoising diffusion model~\cite{rombach2022high}, which operates through a pair of forward and backward Markov chains. 
In the forward process, Gaussian noise $\epsilon_t$ is progressively added to erode $z_t$, according to a pre-defined noise schedule~\cite{karras2022elucidating}.
Conversely, the backward process performs step-by-step denoising, estimating $\epsilon_t$ to recover from $z_{t-1}$ from $z_t$. The backward process is conditioned on the motion feature and content feature as guidance. We refer to~\cite{ho2020denoising} for closed-form equations of the diffusion process.
We use DiT~\cite{peebles2023scalable} as the backbone for our diffusion network. To ensure temporal smoothness in the generated latents, we insert additional temporal attention layers between each DiT spatial block, following~\cite{blattmann2023align}.
We insert the motion condition by adding the motion feature to the diffusion timestep embedding, and insert the content condition by concatenating the content feature with the noisy input.

\subsection{Model Training}
\label{sec:method:training}
\paragraph{\textbf{Loss Functions}}
Following classical rate-distortion approach~\cite{li2021deep}, we train our model using both diffusion denoising loss $\mathcal{L}_d$ and the bitrate loss $\mathcal{L}_{VQ}$ with a weight factor $\lambda$:
\begin{align} 
\label{eq:d_loss}  
    \mathcal{L}_d &= MSE(z, \Tilde{z})  \\
\label{eq:vq_loss}  
    \mathcal{L}_{VQ} &= MSE(\mathcal{H}, \mathcal{\hat{H}}) \\  
\label{eq:total_loss}
    \mathcal{L} &= \mathcal{L}_d + \lambda \mathcal{L}_{VQ}
\end{align}  
We set $\lambda$ to 0.04 in our experiments.

\paragraph{\textbf{Cross-driven Strategy}}
To further avoid collapsing into trivial, entangled representations, each training video clip is evenly divided along the temporal axis into two segments with similar semantic content but distinct motion dynamics.
During training, the content feature from the first segment and the motion feature from the second segment are used to reconstruct the latter segment of the video.


\paragraph{\textbf{Implementation Details}}
We use the pretrained image VAE from Stable Diffusion 2.0~\cite{rombach2022high} to tokenize input frame sequences.
The transformer encoder has 12 layers with a hidden size of 512, a feedforward dimension of 2048 and 8 attention heads. The diffusion decoder is a DiT-B/4.
The low-bitrate codebook has 64 groups, each with 32 codes.
The temperature of the gumbel-softmax operator is annealed from $1.1$ to $0.5$ with a factor of $0.999$. 
We adopt the EDM~\cite{karras2022elucidating} framework for diffusion training and sampling. 
Please see the experiment section for training hyper-parameters and statistics.

%% file: figures/pipeline.tex
\begin{figure*}[tb]
    \centering
    \includegraphics[width=0.95\textwidth]{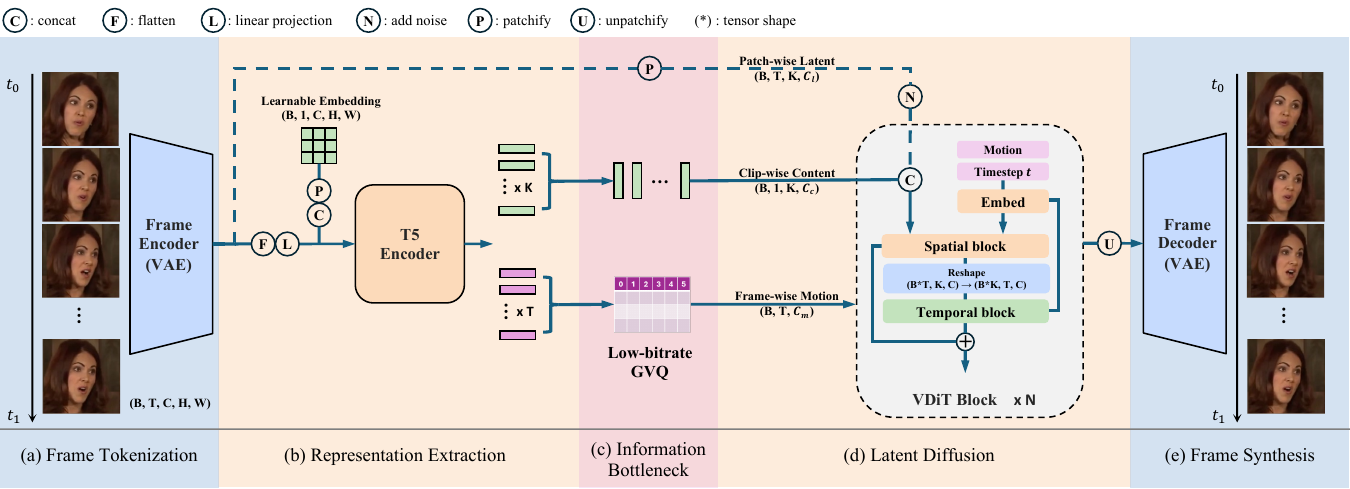}
    \caption{BCD is a diffusion model which consists of (a) a pre-trained image encoder to generate latent features for each video frame; (b) a latent transformer to extract a clip-wise content feature and frame-wise motion features; (c) a low-bitrate vector quantization module serving as information bottleneck on motion features to ensure disentanglement;
    (d) a latent video diffusion decoder using both content and motion features as conditions;
    (e) paired image decoder of (a) to synthesize final video frames.
    }  
    \label{fig:pipeline}
\end{figure*}

%% file: sec/4_experiments.tex
\section{Experiments}
\label{sec:experiments}
To validate our proposed BCD framework, we conduct experiments on two video datasets with distinctly different data distributions.
Our main experiments are conducted on a real world video dataset consists of massive internet collected talking head videos. We choose talking heads as our primary demonstration scenario because it represents a significant real-world application of video disentanglement with in-the-wild video inputs. Additionally, fidelity and disentanglement quality can be clearly assessed through cross-identity motion transfer and video generation quality.
To make our experiments more comprehensive, we also conduct extensive experiment on a pixel-art style 2D character dataset, commonly used in the disentanglement literature, to provide additional disentanglement evaluation as well as to demonstrate the generality of our method.

\subsection{Datasets}
\noindent{\textbf{LRS3~\cite{afouras2018lrs3}}} is a talking head dataset consists of face head crops from over 400 hours of TED and TEDx videos. All videos are converted to 25fps and resized to 256$\times$256.
\noindent{\textbf{Sprites~\cite{reed2015deep}}} is a synthetic cartoon character video dataset. Each character's dynamic motion falls into one of three action categories (walking, spellcasting, or slashing) performed in three possible directions (left, front, and right). The static content includes the character’s skin color, top color, pants color, and hair color, each with six possible variations. Each sequence consists of 8 frames with a resolution of 64$\times$64.

\subsection{Experiment Setup on LRS3}
\paragraph{\textbf{Hyper-parameters}}
We set a target bitrate $H_{target}=160$ which corresponds to 4kbps bitrate with 25fps video.
The video batch size is set to 32 with 50 frames for each video clip.
We train our model for 30 epochs without temporal layer, followed by 15 epochs with temporal layer, using AdamW~\cite{loshchilov2017decoupled} optimizer. 
The learning rate for first 30 epochs is $0.0001$ for the decoder and $0.0002$ for the encoder. For final 15 epochs, we reduce the learning rate of the encoder by half.
Training on LRS3 takes $\sim$ 4–5 days on 8xA100 GPUs with 80GB memory, with an inference time of $\sim$ 30 seconds for a 50-frame video on a single A100.
\paragraph{\textbf{Metrics}}
For the LRS3 dataset, we adopt the FID score~\cite{heusel2017gans} and the cosine similarity (CSIM) computed from face recognition networks~\cite{deng2018arcface} to evaluate the (perceptual) video fidelity following previous talking head works.
To evaluate disentanglement, 
we use a 3D head reconstruction method~\cite{wood20223d, hewitt2024look} to fit parametric mesh coefficients and reconstruct identity mesh, motion mesh, and cross-driven ground truth mesh for reference and generated videos.
We then measure the identity error, motion error, and cross-driven error by computing the mean-squared vertex distance between corresponding mesh pairs.
Please refer to the Appendix for more details.
\paragraph{\textbf{Baselines}} For talking head tasks, there exists a bunch of existing works using different types of priors.
We chose the following methods representing typical types of priors as our baseline: FOMM~\cite{siarohin2019first}, MCNET~\cite{hong23implicit} and LivePortrait~\cite{guo2024liveportrait} with learned keypoints and piece-wise affine flow,
HyperReenact~\cite{bounareli2023hyperreenact} with 3D parametric face priors, and LIA~\cite{wang2021latent} with linear orthogonal motion basis.

\subsection{Main Results on Talking Heads}

\input{figures/result_extreme_pose}

\input{figures/motion_generation}

\input{tables/results_cross}
\paragraph{\textbf{Motion Transfer}}
The motion transfer result is shown quantitatively in \cref{tab:cross_identity}.
Although our method does not use any human face related prior,
it performs the best in terms of identity error, motion error, cross transfer error, and FID score; we also achieve a reasonable score of CSIM. We have attached more results in the Appendix.

To further compare our method to other baselines with different types of inductive bias, we conduct an analysis on results in the test set, and visualize generated frames in \cref{fig:extreme_pose}. 
FOMM~\cite{siarohin2019first}, MCNET~\cite{hong23implicit} and LivePortrait~\cite{guo2024liveportrait} share similarities with the usage of self-supervised key-point as motion representations, which employs relative key-point locations to account for identity preservation with higher CSIM. 
However, they have significant larger motion (and identity) errors due to warping artifacts with large motion as well as its \textit{same pose assumptions} between the reference frame and the first frame of the driving video. 
LIA~\cite{wang2021latent} has the largest motion error because it exhibits a very constrained motion space with its linear combination of motion basis. Visualizations demonstrated that generated frames have good identity preservation but often completely fail to transfer motion.
HyperReenact~\cite{bounareli2023hyperreenact} leverages a 3D face prior, enabling more accurate 3D vertex positions and thus lower motion error. However, the explicit usage of a 3DMM~\cite{deng2019accurate} face prior makes their generated frames look like 3DMM renderings. This is reflected in its worst FID scores and can be clearly observed from the qualitative results.
Compared with all these methods using different types of inductive bias, our method can synthesize high-quality motion transfer results under diverse motion inputs, even for some extreme motions while all other methods failed (\cref{fig:extreme_pose}, the first row).
Overall, these results clearly demonstrate that our method is able to synthesize high-fidelity videos for motion transfer via proper disentanglement of motion and content.

\paragraph{User Study.} In addition to quantitative evaluation, we conduct a user study for subjective evaluation.
The study consists of 15 groups of video reference and generated baseline videos. Users are then asked to score each generated video in terms of (1) identity preservation, (2) motion consistency and (3) visual quality.
We collect responses from 18 participants in total.
The result is shown in \cref{tab:user_study}. As shown, we have achieved the highest score compared to other baselines, demonstrating that our disentanglement also matches human perception of video data.

\input{tables/user_study}

\paragraph{\textbf{Video Generation}}
Our learned video space provides a structured motion representation for video generation. We demonstrate this by training an auto-regressive transformer directly on our motion spaces for the video generation task.
Specifically, we train a GPT-2~\cite{radford2019language} model on motion features generated by the BCD model on the talking head data. The content feature is inserted at the beginning as the prompt.
\Cref{fig:motion_generation} visualizes some of our generated frames. We are able to generate video sequences with reasonable motion patterns based on a single-frame prompt. This experiment confirms that our motion space captures the full motion distribution rather than memorizing data points in the training set, enabling content-preserved motion generation via sampling.
We refer to the Appendix for the training details.

\subsection{Discussions}
In this section, we discuss some of our design considerations, limitations, and societal impact.

\input{tables/ablation_bitrate}
\paragraph{\textbf{Bitrate}}
The most important hyperparameter for BCD is the target bitrate during training.
Excessive bitrate does not serve as an effective information bottleneck, resulting in diminished decoupling capabilities; conversely, a bitrate that is too low leads to insufficient transmission of information, making it struggle to synthesize high-quality videos. Both cases would lead to an increase in error for motion transfer.
We empirically set the target bitrate for motion to 4kbps, slightly lower than a recent method~\cite{gao2024implicit} that reports an average bitrate of approximately 5kbps for talking head videos, to promote motion disentanglement.
We validate our bitrate selection in \cref{tab:ablation_bitrate}, which presents talking head motion transfer results across target bitrates ranging from 2 kbps (80 bits per frame) to 8 kbps (320 bits per frame).
We observed that the minimum cross-driven error occurs at approximately 4kbps.

\paragraph{Input sequence lengths} 
Our video disentanglement model supports a flexible number of input frames for both motion and content.
As shown in \cref{tab:ablation_bitrate} (row ``4.0-single”), our method remains robust even with a single content frame, enabling one-shot input. Our method can be integrated with common techniques for handling transformers with long contexts, such as sliding windows. 

\paragraph{Training strategy} The cross-driven strategy (\cref{sec:method:training}) also benefits our disentanglement. The last row of \cref{tab:ablation_bitrate} shows that disabling this strategy during training results in increased motion transfer error.


\paragraph{\textbf{Societal impact}}
Our research focuses on the self-supervised decoupling of motion and content from video. 
While we present results on human faces, our goal is solely to demonstrate our disentanglement capability on real videos. We have no intention of generating content for misleading or deceptive purposes.
We are opposed to any behavior of creating misleading or harmful contents of real persons, and we are also interested in applying our technique for advancing forgery detection.
\paragraph{\textbf{Limitations}}
Our method requires substantial training data to disentangle videos with minimal task-specific inductive bias. 
The optimal bitrate for BCD to achieve the best disentangle can be dataset dependent.
We are still observing a slight amount of video flickering even after the temporal fine-tuning.
Real-world video datasets typically contain more dynamic variations and fewer static variations. This may lead to a degradation in the model's performance to capture static elements when encountering completely out-of-distribution video inputs.

\input{figures/motion_transfer_lpc}
\subsection{Results on Sprites dataset}
Our proposed BCD framework is not restricted to talking head video data only, and it can potentially be applied to other video datasets that have significant different distributions.
We demonstrate the generality of our BCD
framework by training our model on the LPC Sprites dataset, which has been widely adopted in the disentanglement research field~\cite{li2018disentangled, bai2021contrastively, zhu2020s3vae, simon2024sequential}
.
The LPC Sprites dataset consists of pixel-art style, cartoon character videos in a lower resolution (4× lower than talking head videos) and fewer video clips, with a more artistic style. Hence we reduce the target bitrate $H_{target}$ to 6 (correspond to 150bps), and directly training in pixel space instead of pre-trained image latent space.
\Cref{fig:motion_transfer_lpc} demonstrates motion transfer results on the LPC Sprites test dataset. 
Our motion transfer results demonstrate superior attribute accuracy. See the Appendix for details.



%% file: figures/result_extreme_pose.tex
\begin{figure*}[t]
\centering
\begin{minipage}{0.12\linewidth}
\centering
\scriptsize Content Ref.
\end{minipage}
\begin{minipage}{0.12\linewidth}
\centering
\scriptsize Motion Ref.
\end{minipage}
\begin{minipage}{0.12\linewidth}
\centering
\scriptsize FOMM
\end{minipage}
\begin{minipage}{0.12\linewidth}
\centering
\scriptsize MCNET
\end{minipage}
\begin{minipage}{0.12\linewidth}
\centering
\scriptsize HyperReenact
\end{minipage}
\begin{minipage}{0.12\linewidth}
\centering
\scriptsize LIA
\end{minipage}
\begin{minipage}{0.12\linewidth}
\centering
\scriptsize LivePortrait
\end{minipage}
\begin{minipage}{0.12\linewidth}
\centering
\scriptsize \textbf{Ours}
\end{minipage}
\\
\begin{minipage}{0.12\linewidth}
\includegraphics[width=\textwidth]{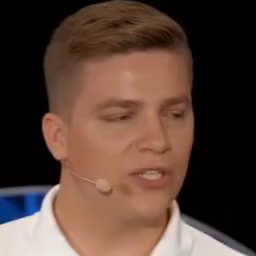} 
\end{minipage}
\begin{minipage}{0.12\linewidth}
\includegraphics[width=\textwidth]{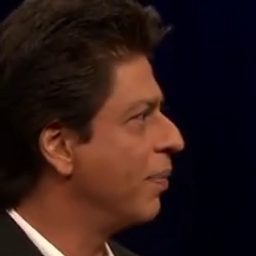} 
\end{minipage}
\begin{minipage}{0.12\linewidth}
\includegraphics[width=\textwidth]{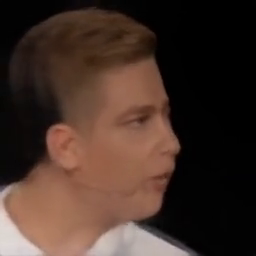} 
\end{minipage}
\begin{minipage}{0.12\linewidth}
\includegraphics[width=\textwidth]{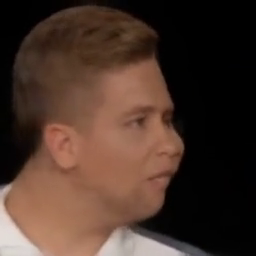} 
\end{minipage}
\begin{minipage}{0.12\linewidth}
\includegraphics[width=\textwidth]{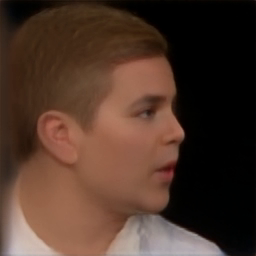} 
\end{minipage}
\begin{minipage}{0.12\linewidth}
\includegraphics[width=\textwidth]{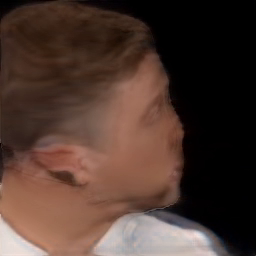} 
\end{minipage}
\begin{minipage}{0.12\linewidth}
\includegraphics[width=\textwidth]{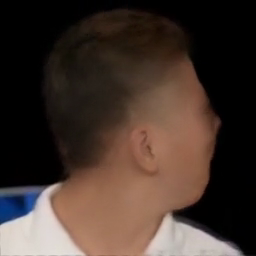} 
\end{minipage}
\begin{minipage}{0.12\linewidth}
\includegraphics[width=\textwidth]{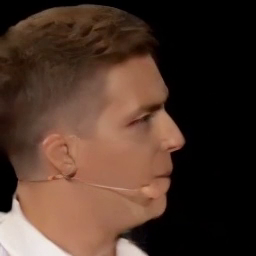} 
\end{minipage}
\\
\begin{minipage}{0.12\linewidth}
\includegraphics[width=\textwidth]{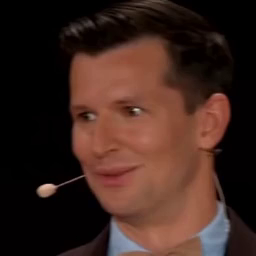} 
\end{minipage}
\begin{minipage}{0.12\linewidth}
\includegraphics[width=\textwidth]{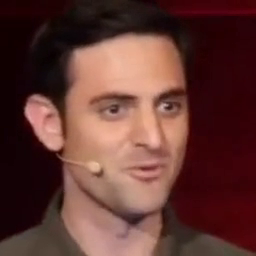} 
\end{minipage}
\begin{minipage}{0.12\linewidth}
\includegraphics[width=\textwidth]{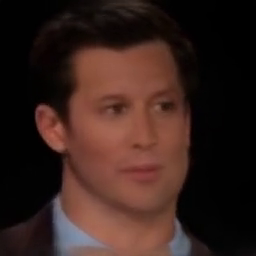} 
\end{minipage}
\begin{minipage}{0.12\linewidth}
\includegraphics[width=\textwidth]{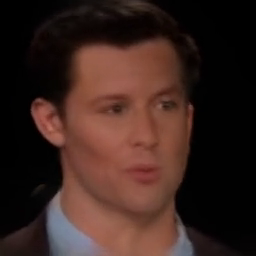} 
\end{minipage}
\begin{minipage}{0.12\linewidth}
\includegraphics[width=\textwidth]{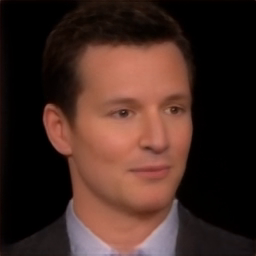} 
\end{minipage}
\begin{minipage}{0.12\linewidth}
\includegraphics[width=\textwidth]{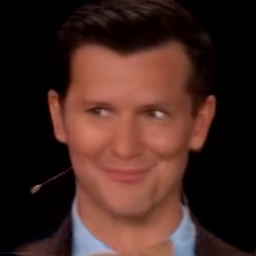} 
\end{minipage}
\begin{minipage}{0.12\linewidth}
\includegraphics[width=\textwidth]{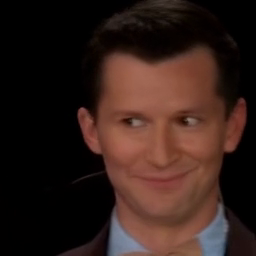} 
\end{minipage}
\begin{minipage}{0.12\linewidth}
\includegraphics[width=\textwidth]{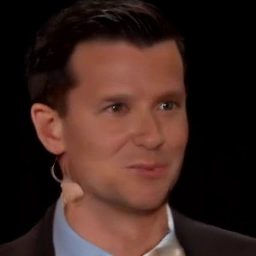} 
\end{minipage}
\\
\begin{minipage}{0.12\linewidth}
\includegraphics[width=\textwidth]{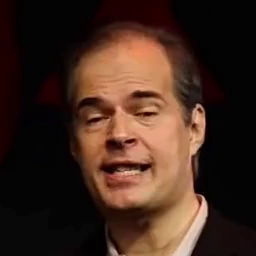} 
\end{minipage}
\begin{minipage}{0.12\linewidth}
\includegraphics[width=\textwidth]{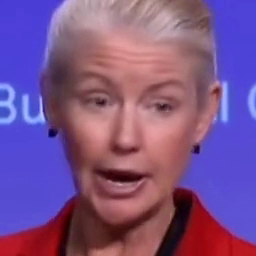} 
\end{minipage}
\begin{minipage}{0.12\linewidth}
\includegraphics[width=\textwidth]{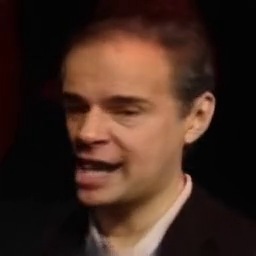} 
\end{minipage}
\begin{minipage}{0.12\linewidth}
\includegraphics[width=\textwidth]{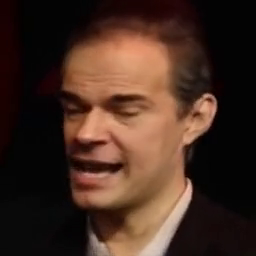} 
\end{minipage}
\begin{minipage}{0.12\linewidth}
\includegraphics[width=\textwidth]{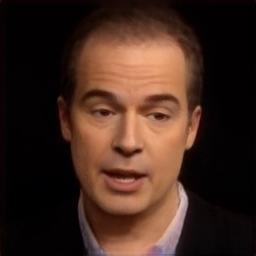} 
\end{minipage}
\begin{minipage}{0.12\linewidth}
\includegraphics[width=\textwidth]{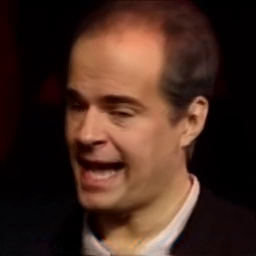} 
\end{minipage}
\begin{minipage}{0.12\linewidth}
\includegraphics[width=\textwidth]{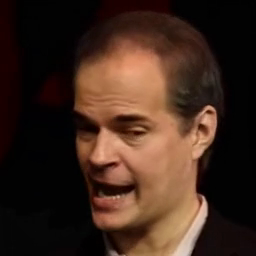} 
\end{minipage}
\begin{minipage}{0.12\linewidth}
\includegraphics[width=\textwidth]{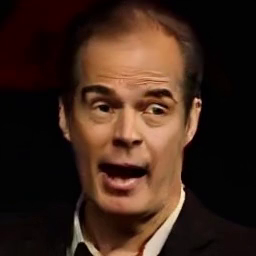} 
\end{minipage}
\\
\begin{minipage}{0.12\linewidth}
\includegraphics[width=\textwidth]{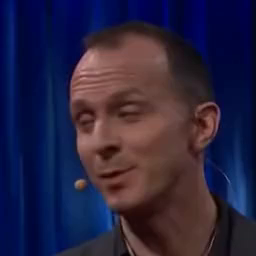} 
\end{minipage}
\begin{minipage}{0.12\linewidth}
\includegraphics[width=\textwidth]{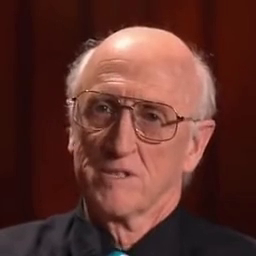} 
\end{minipage}
\begin{minipage}{0.12\linewidth}
\includegraphics[width=\textwidth]{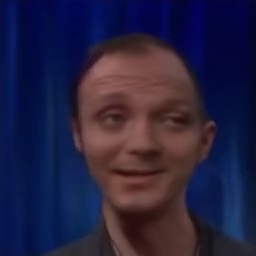} 
\end{minipage}
\begin{minipage}{0.12\linewidth}
\includegraphics[width=\textwidth]{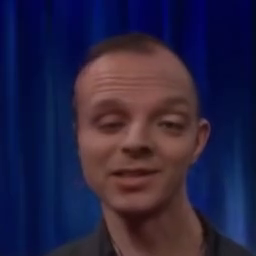} 
\end{minipage}
\begin{minipage}{0.12\linewidth}
\includegraphics[width=\textwidth]{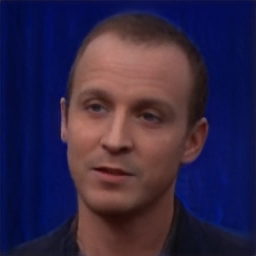} 
\end{minipage}
\begin{minipage}{0.12\linewidth}
\includegraphics[width=\textwidth]{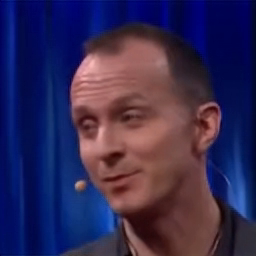} 
\end{minipage}
\begin{minipage}{0.12\linewidth}
\includegraphics[width=\textwidth]{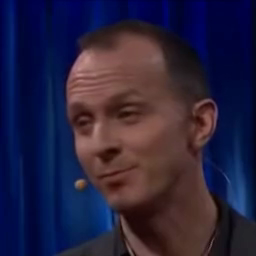} 
\end{minipage}
\begin{minipage}{0.12\linewidth}
\includegraphics[width=\textwidth]{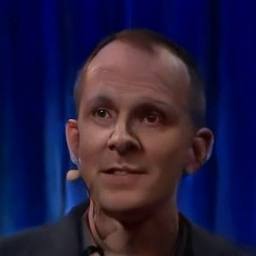} 
\end{minipage}
\\
\begin{minipage}{0.12\linewidth}
\includegraphics[width=\textwidth]{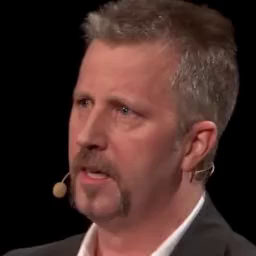} 
\end{minipage}
\begin{minipage}{0.12\linewidth}
\includegraphics[width=\textwidth]{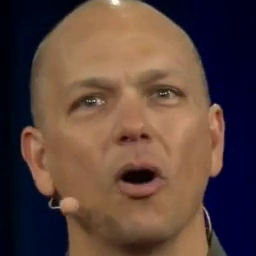} 
\end{minipage}
\begin{minipage}{0.12\linewidth}
\includegraphics[width=\textwidth]{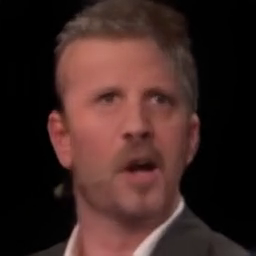} 
\end{minipage}
\begin{minipage}{0.12\linewidth}
\includegraphics[width=\textwidth]{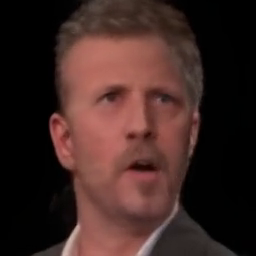} 
\end{minipage}
\begin{minipage}{0.12\linewidth}
\includegraphics[width=\textwidth]{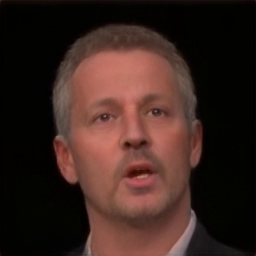} 
\end{minipage}
\begin{minipage}{0.12\linewidth}
\includegraphics[width=\textwidth]{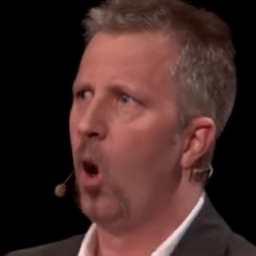} 
\end{minipage}
\begin{minipage}{0.12\linewidth}
\includegraphics[width=\textwidth]{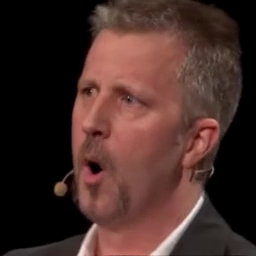} 
\end{minipage}
\begin{minipage}{0.12\linewidth}
\includegraphics[width=\textwidth]{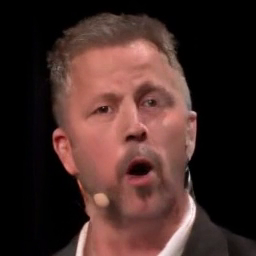} 
\end{minipage}

\caption{Motion transfer results on different input videos. The reference appearance and motion frames are shown in the first two colums.}
\label{fig:extreme_pose}
\end{figure*}

%% file: figures/motion_generation.tex
\begin{figure*}[tbh]
    \centering
    \begin{minipage}[b]{0.16\linewidth}
        \centering
        \scriptsize Prompt
        \\
        \includegraphics[width=0.95\textwidth]{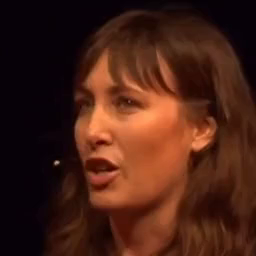}
        \\
        \includegraphics[width=0.95\textwidth]{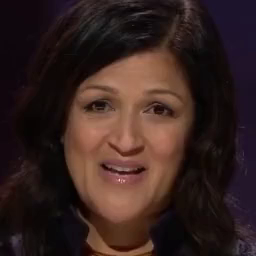}
    \end{minipage}
    \tikz[overlay, remember picture] \draw[dashed,thick] ([xshift=0.302\linewidth]current page.west|-0,0.233\textheight) -- ([xshift=0.302\linewidth]current page.west|-0,0);
    \begin{minipage}[b]{0.80\linewidth}
        \centering
        \scriptsize Generated Frames
        \\
        \includegraphics[width=0.19\textwidth]{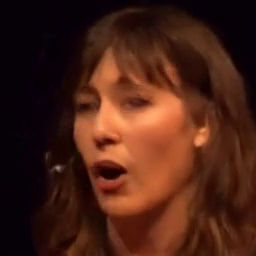}
        \includegraphics[width=0.19\textwidth]{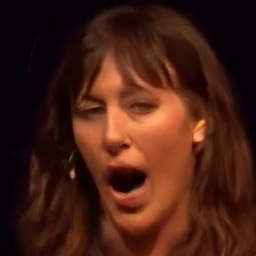}
        \includegraphics[width=0.19\textwidth]{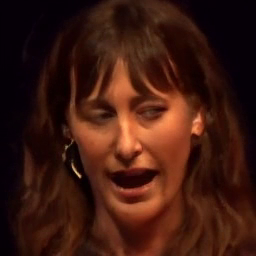}
        \includegraphics[width=0.19\textwidth]{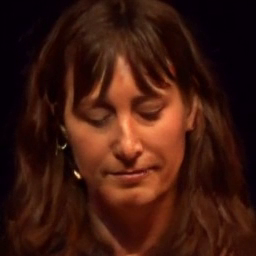}
        \includegraphics[width=0.19\textwidth]{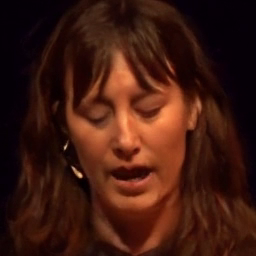}
        \\
        \includegraphics[width=0.19\textwidth]{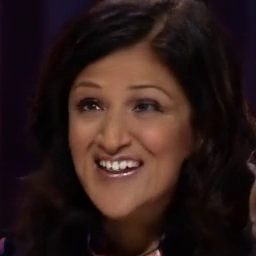}
        \includegraphics[width=0.19\textwidth]{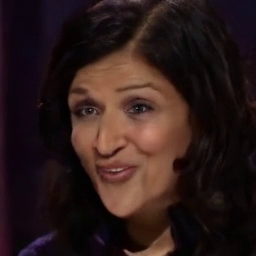}
        \includegraphics[width=0.19\textwidth]{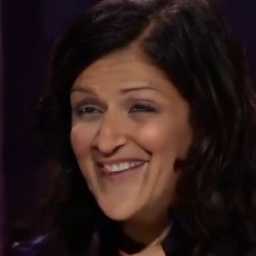}
        \includegraphics[width=0.19\textwidth]{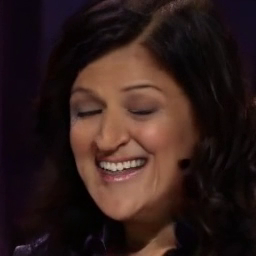}
        \includegraphics[width=0.19\textwidth]{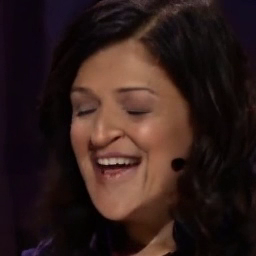}

    \end{minipage}
    \caption{Motion generation results. We use the content and motion feature of a single frame as the prompt shown on the left; five of the generated frames are shown on the right.}
    \label{fig:motion_generation}
\end{figure*}

%% file: tables/results_cross.tex
\begin{table}[htb]
\centering
\caption{Comparisons on cross identity motion transfer. Best and runner-up methods are marked with bold and underlines.}
\label{tab:cross_identity}
\scalebox{0.7}{ 
\begin{tabular}{@{}cccccc@{}}
\toprule
Method             & FID$\downarrow$ & CSIM$\uparrow$ & \begin{tabular}[c]{@{}c@{}}Identity error$\downarrow$\\ $\times 10^{-1}$\end{tabular} & \begin{tabular}[c]{@{}c@{}}Motion error$\downarrow$\\ $\times 10^{-2}$\end{tabular} & \begin{tabular}[c]{@{}c@{}}Cross error$\downarrow$\\ $\times 10^{-2}$\end{tabular} \\ \midrule
FOMM          & {\ul 98.5} & \textbf{0.76} &0.75           & 24.3       & 24.1               \\
MCNET         & 98.6 & \textbf{0.76} & 0.85          & 23.9       & 23.6               \\
HyperReenact      & 106.8 & 0.58 & {\ul 0.57}          & {\ul 3.94}       & {\ul 4.68}              \\
LIA               & 104.4 & {\ul 0.71} & {\ul 0.57}           & 36.1       & 34.1               \\
LivePortrait               & 100.3 & 0.69 & 0.66           & 24.6       & 23.7  \\ \midrule
Ours              & \textbf{86.0} & 0.69 & \textbf{0.41} & \textbf{3.13} & \textbf{3.67}      \\
\bottomrule
\end{tabular}
}
\end{table}

%% file: tables/user_study.tex

\begin{table}[htb]
\centering
\caption{User study on identity preservation, motion consistency and visual quality. Best and runner-up methods are marked with bold and underlines.}
\label{tab:user_study}
\scalebox{0.68}{ 
\begin{tabular}{@{}cccc@{}}
\toprule
Method & Identity Preservation $\uparrow$  & Motion Consistency $\uparrow$ & Visual Quality $\uparrow$ \\ \midrule
FOMM      & 3.34 & 2.63 &  2.84  \\
MCNET     & 3.36 & 2.74 &  2.94  \\
HyperReenact & 2.97 & {\ul 4.01} &  {\ul 3.72}   \\
LIA & {\ul 3.66} & 2.53 &  3.53   \\
\midrule
Ours & \textbf{4.10} & \textbf{4.30} &  \textbf{4.00}   \\
 \bottomrule
\end{tabular}
}
\end{table}

%% file: tables/ablation_bitrate.tex
\begin{table}[htb]
\centering
\caption{Quantitative results under different training and evaluation setup. Best and runner-up methods are marked with bold and underlines.}
\label{tab:ablation_bitrate}
\scalebox{0.68}{ 
\begin{tabular}{@{}cccccc@{}}
\toprule
\begin{tabular}[c]{@{}c@{}}Target Bitrate\\ (kbps)\end{tabular} & FID$\downarrow$ & CSIM$\uparrow$ & \begin{tabular}[c]{@{}c@{}}Shape error$\downarrow$\\ $\times 10^{-1}$\end{tabular} & \begin{tabular}[c]{@{}c@{}}Motion error$\downarrow$\\ $\times 10^{-2}$\end{tabular} & \begin{tabular}[c]{@{}c@{}}Cross error$\downarrow$\\ $\times 10^{-2}$\end{tabular} \\ \midrule
2.0          & 88.5 & \textbf{0.71} & \textbf{0.34}           & 5.26       & 5.68               \\
6.0      & {\ul 87.6} & 0.68 & 0.56          & 3.23       & 4.13               \\
8.0               & 89.3 & 0.66 & 0.49          & \textbf{3.04}       & {\ul 3.74}               \\
\midrule
4.0         & \textbf{86.0} & {\ul 0.69} & {\ul 0.41}          & {\ul 3.13}       & \textbf{3.67}     \\
\midrule
\begin{tabular}[c]{@{}c@{}}4.0\\ (single ref.)\end{tabular} & 87.9 & {\ul 0.69} & 0.47    & {\ul 3.13}  &  3.81         \\
\midrule
\begin{tabular}[c]{@{}c@{}}4.0\\ (w/o. cross-driven)\end{tabular} & 120.1 & 0.64 & 0.58    & 41.5 &  40.4         \\ 
 \bottomrule
\end{tabular}
}
\end{table}

%% file: figures/motion_transfer_lpc.tex
\begin{figure}[t]
\centering
\begin{minipage}{0.48\linewidth}
\centering
\tiny Input A
\end{minipage}
\begin{minipage}{0.48\linewidth}
\centering
\tiny Input B
\end{minipage}
\\
\begin{minipage}{0.48\linewidth}
\includegraphics[width=\textwidth]{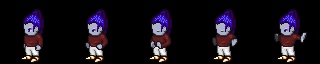} 
\end{minipage}
\begin{minipage}{0.48\linewidth}
\includegraphics[width=\textwidth]{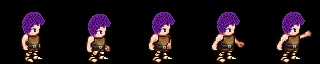} 
\end{minipage}
\\
\begin{minipage}{0.48\linewidth}
\centering
\tiny motion of A + content of B
\end{minipage}
\begin{minipage}{0.48\linewidth}
\centering
\tiny motion of B + content of A
\end{minipage}
\\
\begin{minipage}{0.48\linewidth}
\includegraphics[width=\textwidth]{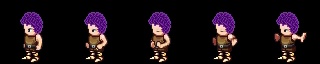} 
\end{minipage}
\begin{minipage}{0.48\linewidth}
\includegraphics[width=\textwidth]{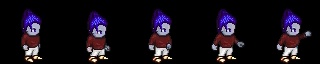} 
\end{minipage}
\\
\caption{Motion transfer results on Sprites test set.}
\label{fig:motion_transfer_lpc}
\end{figure}

%% file: sec/5_conclusion.tex
\section{Conclusion}
\label{sec:conclusion}
We have proposed Bitrate Controlled Diffusion (BCD) for learning disentangled video representations into motion and semantic contents.
BCD represents motion and contents with less inductive bias, leveraging the information bottleneck with low-bitrate vector quantization to effectively disentangle them in latent space.
The training process is self-supervised with a denoising diffusion task.
We have demonstrated the fidelity and the disentanglement of our method on real world talking head videos with motion transfer and generation tasks. We have also illustrated the generalization ability of our method by demonstrate results on both real world video data and synthetic cartoon video data.
We believe our work offers a new perspective on disentangled video representation learning.
\paragraph{Future work.} Future avenues include expanding our method to general video data with broader applications, exploring the interactivity of the learned motion latent space through post-hoc mapping, as well as establishing a more theoretical elaboration for the optimal bitrate to achieve disentanglement.

%% file: sec/7_acknowledgement.tex
\paragraph{Acknowledgements}
The authors would like to thank Yue Gao for discussion, Tadas Baltrusaitis, Charlie Hewitt and Paul Mcllroy for providing 3D mesh fitting packages.
Xie Chen is supported by the Shanghai Municipal Science and Technology Major Project under Grant 2021SHZDZX0102 and Yangtze River Delta Science and Technology Innovation Community Joint Research Project (Grant No. 2024CSJGG01100).

%% file: sec/6_suppl.tex
\appendix
\label{appendix}
\section*{Appendix}
The appendix contains additional implementation details for our training and evaluation as well as more results and ablations.

\section{Implementation Details}
\subsection{Video Representation Learning}
\paragraph{Diffusion Model Preliminaries}
Diffusion models~\cite{ho2020denoising} have emerged as a powerful class of generative models that achieve state-of-the-art performance in various generation tasks. These models operate by defining a stochastic Markov process that progressively transforms data into noise and then learns to reverse this process to generate realistic samples.
Formally, a diffusion model consists of a forward process (a.k.a., the diffusion process) and a reverse process (a.k.a., the denoising process). The forward process is defined as a Markov chain that iteratively adds Gaussian noise to a data sample $\mathbf{z}_0 \sim q(\mathbf{z})$ over $T$ steps\footnote{We omit the video timestep here for simplicity. The subscript $t$ denoted the denoising timestep.}:
\begin{equation}
\label{eq:appendix:diffusion_forward}
    q(\mathbf{z}_t | \mathbf{z}_{t-1}) = \mathcal{N}(\mathbf{z}_t; \sqrt{\alpha_t} \mathbf{z}_{t-1}, (1 - \alpha_t) \mathbf{I}),
\end{equation}
where $\{\alpha_t\}$ are variance schedule parameters controlling the noise level at each step. By applying this process iteratively, the data distribution gradually approaches an isotropic Gaussian distribution.
The reverse process, parameterized by a neural network $\boldsymbol{\theta}$, aims to learn the conditional distribution:
\begin{equation}
\label{eq:appendix:diffusion_backward}
    p_{\boldsymbol{\theta}}(\mathbf{z}_{t-1} | \mathbf{z}_t, \mathbf{C}) = \mathcal{N}(\mathbf{z}_{t-1}; \boldsymbol{\mu}_{\boldsymbol{\theta}}(\mathbf{z}_t, t, \mathbf{C}), \boldsymbol{\Sigma}_{\boldsymbol{\theta}}(\mathbf{z}_t, t, \mathbf{C})).
\end{equation}
where $C$ is the condition signal to guide the denoising process. In our case, the conditional signal comes from our disentangle encoder $\mathcal{T}$, i.e., $C=(m, c)=T(z_0)$.
\cite{ho2020denoising} has revealed that by doing simplification and using parametrization trick, we can derive a closed form estimation for $z_0$ from \cref{eq:appendix:diffusion_forward} and \cref{eq:appendix:diffusion_backward}:
\begin{equation}
    \mathbf{z}_0 \approx \frac{1}{\sqrt{\bar{\alpha}_t}} (\mathbf{z}_t - \sqrt{1 - \bar{\alpha}_t} \boldsymbol{\epsilon}_{\boldsymbol{\theta}}(\mathbf{z}_t, t, \mathbf{C})),
\end{equation}
where $\bar{\alpha}_t = \prod_{s=1}^{t} \alpha_s$. Sampling is performed by iteratively denoising from $\mathbf{x}_T \sim \mathcal{N}(\mathbf{0}, \mathbf{I})$ back to $\mathbf{x}_0$.

The added gaussian noise $\boldsymbol{\epsilon}_{\boldsymbol{\theta}}(\mathbf{x}_t, t, \mathbf{C})$ is predicted by the denoising decoder $\mathcal{D}$. 

The model is trained by predicting the noise $\boldsymbol{\epsilon}$ added at each step. Specifically, the training loss is formulated as:
\begin{equation}
    \mathcal{L}(\boldsymbol{\theta}) = \mathbb{E}_{\mathbf{x}_0, \boldsymbol{\epsilon}, t} \Big[ \|\boldsymbol{\epsilon} - \boldsymbol{\epsilon}_{\boldsymbol{\theta}}(\mathbf{x}_t, t)\|_2^2 \Big],
\end{equation}
where $\mathbf{x}_t$ is obtained by perturbing $\mathbf{x}_0$ with noise $\boldsymbol{\epsilon} \sim \mathcal{N}(\mathbf{0}, \mathbf{I})$ according to the forward process. This training objective effectively reduces to a denoising autoencoder loss, enabling the model to learn an implicit score function that guides the reverse process.

\paragraph{Data pre-processing.}
We further apply a 2x temporal down-sampling to all video data during training.
This is mainly to increase the maximum motion range the model can see during training under limited computational resources.
All input video sequences are cropped and split into 4-second clips.
Our cross-driven strategy (see main text) of splitting a clip in half assumes minimal content changes within each training clip, which can be ensured through data preprocessing.

\paragraph{Model Architecture and Hyper-parameters.}
For our disentangle encoder which extracts motion and content simultaneously, we adopt the T5 Encoder equipped with relative positional encoding. For diffusion denoiser, we use DiT architecture and insert temporal blocks in certain layers (as listed in temporal block indices of \cref{tab:hyperparameters_representation} and \cref{tab:hyperparameters_representation_sprites}), the temporal blocks and spatial blocks follow the original design of DiT blocks \cite{peebles2023scalable}.
The motion feature $m$ is inserted into the decoder $D$ by modulating the activations of each layer h using the Adaptive Group Normalization layers in each DiT block:
\begin{equation}
    AdaGN(h, m, t) = m_s(t_sGroupNorm(h) + t_b) + m_b
\end{equation}
 where $(t_s,t_b)$ and $(z_s,z_b)$ are obtained by linear projections from the sinusoidal timestep embedding of denoising timestep t and the motion feature $m$ respectively.
 The content feature $c$ is directly concatenated to the input $z_t$.

Hyper-parameters for our representation learning model are listed in \cref{tab:hyperparameters_representation} and \cref{tab:hyperparameters_representation_sprites}. 
Given the distinct data domains and dataset scales between the talking head dataset LRS3 and the synthetic cartoon characters dataset Sprites, we employ different hyperparameters for the respective models. Additionally, instead of utilizing a pre-trained VAE in the Sprites model, we directly process the video in the pixel space.

\subsection{Auto-regressive motion generation}
\paragraph{Data processing.}
Our auto-regressive motion generation model is trained on sequences of motion tokens; we generate these tokens from our pre-trained BCD model. The clip-wise content token is prepended to motion token sequences as the first sequence element, serving as a content prompt.
To increase the motion diversity for both large motion and subtle motion, we implement data augmentation by downsampling videos temporally by a factor of 2 with a 50\% probability.

\paragraph{Implementation.}
We use the GPT2 implementation from huggingface~\cite{wolf2019huggingface}.
We modify the output head of GPT2 to match our grouped codebook outputs, i.e., we output the probability distribution of the next motion token across multiple codebooks, resulting in multi-group probability logits.
We calculate the cross entropy loss between these probability distributions for each codebook with the ground truth motion tokens (in terms of codebook IDs) of the next timestep. 
The average of these cross-entropy losses serves as the loss function for our model.
For inference, we sample predicted logits with the largest probability.
Output video frames are generated by running our diffusion denoiser with predicted motion sequences and content features.
Hyper-parameters for our motion generation model are listed in \cref{tab:hyperparameters_gpt}.

\section{Additional Results}

\input{figures/supp_sprites}
\subsection{Motion Transfer and Generation}
We show more motion transfer results on Sprites in \cref{fig:supp_sprites}. Following previous methods, we also quantitatively assess disentanglement quality by measuring the attribute accuracy of videos generated by our model, using the pre-trained video classification network from~\cite{bai2021contrastively}. \cref{tab:results_sprites} reports the motion and content accuracy for each attribute category, demonstrating that our method produces correct disentanglement results and achieves state-of-the-art performance on par with C-DSVAE~\cite{bai2021contrastively}.

\input{tables/results_sprites}

\input{figures/ablation_bitrate}
\input{tables/ablation_content}
\input{tables/ablation_rebuttal}
\subsection{Additional Ablation Studies}
\paragraph{Number of frames for content extraction.}
Our method is able to extract content information from an arbitrary length of video. We analyzed the result of extracting content with different lengths of input and the results are shown in 
\cref{tab:ablation_content}.
Overall, the overall motion transfer quality is quite robust to the number of content frames with slight differences in terms of error metrics, while using more content reference frames improves the identity preservation and cross-identity motion transfer effect.
\paragraph{Effects of different bitrate.}
In the main paper, we quantitatively analyzed the effect of different bitrate. We additionally qualitatively visualize the effect in \Cref{fig:ablation_bitrate}: results under a lower bitrate (e.g. 2kbps) leads to degradation of some motion, while results under larger bitrate ($>$8kbps) demonstrate worse disentanglement of identity.

\paragraph{The usage of bitrate loss.}
We introduced bitrate loss during training to improve bitrate convergence. Ablation studies (see \cref{tab:ablation_rebuttal}) show that removing it degrades performance, confirming its effectiveness.

\section{Evaluation Metrics}
To provide a more thorough evaluation of our method and other baselines on the cross-identity motion transfer task, we proposed additional metrics. These metrics are used together with existing metrics such as CSIM for evaluation results in the main paper. Here we explain the motivation for these additional metrics and their implementation details.
\paragraph{3D Face Fitting.} 
All additional metrics rely on a dense reconstruction of 3D face meshes from both target and input videos. For this purpose, we utilize the method described in Wood et al.\cite{wood20223d, hewitt2024look} for several reasons.
Firstly, it employs an optimization-based approach using dense landmark observations, ensuring accurate face reconstructions.
In contrast, single-pass feed-forward reconstruction methods like those in Deng et al.\cite{deng2019accurate} lack a per-video optimization process and fail to guarantee accurate reconstructions across videos with significant head pose and motion variations.
Secondly, the parametric face model in \cite{wood20223d} incorporates detailed blendshape expression bases within a continuous space, facilitating the assessment of expression transfer. Lastly, the dense landmark predictions and the parametric 3D face model from Wood et al.~\cite{wood20223d} are trained using high-quality synthetic face images covering a wide range of identities, expressions, head poses, genders, and races, minimizing potential ethical biases and privacy concerns during evaluation. In practice, the average fitting error of the face reconstruction method \cite{wood20223d}, as measured by reprojection error in pixel space on our test videos, is less than 1.5 pixels—indicative of high reconstruction quality.

The 3D face reconstruction of a given input video sequence $V$ with $n$ frames is characterized by identity coefficients $\beta \in R^{c}$, per-frame expression coefficients $\psi \in R^{n \times m}$, per-frame pose vector (in Euler angle) $\theta \in R^{n \times 3}$, and per-frame translation vector $t \in R^{n \times 3}$. 
The computation of 3D mesh vertices from these coefficients leverages the parametric mesh model described in \cite{wood2021fake}: $\mathcal{M}(\beta, \theta, \psi, t): R^{c \times m \times 3} \rightarrow R^{v \times 3}$.
For a comprehensive mathematical explanation, we direct readers to the work of \cite{wood2021fake, wood20223d}.
\paragraph{Shape Error.}
The shape error $e_s$ is calculated as the mean squared error (MSE) distance between the \textbf{identity meshes} of the content reference video and the target video. 
An \textbf{identity mesh} $M_{id}$ is derived by setting the pose $\theta$, expression $\psi$, and translation $t$ to zero, i.e.,
\begin{gather}
    M_{identity} = \mathcal{M}(\beta, 0, 0, 0) \\
    e_s = MSE(M^{ref}_{identity}, M^{dst}_{identity})
\end{gather}
Thus, the shape error $e_s$ isolates and quantifies the identity discrepancy between the target and the reference input, excluding all other variables.
\paragraph{Motion Error.}
Likewise, the motion error $e_m$ is determined by calculating the average MSE distance between \textbf{motion meshes} of the motion reference video and the target video.
This calculation is achieved by setting the identity coefficients to zero:
\begin{gather}
    M_{motion} = \mathcal{M}(0, \theta, \phi, t) \\
    e_m = MSE(M^{ref}_{motion}, M^{dst}_{motion})
\end{gather}
It is important to note that head poses are also incorporated into the motion mesh calculation.

\paragraph{Cross Transfer Error.}
Finally, the cross transfer error $e_{c}$ is calculated to evaluate the overall quality of the cross-identity motion transfer task by utilizing the fully fitted meshes:
\begin{gather}
    M_{full} = \mathcal{M}(\beta, \theta, \phi, t) \\
    e_{c} = MSE(M^{ref}_{full}, M^{dst}_{full})
\end{gather}

\input{figures/csim}
\paragraph{Discussion.}
Existing methods evaluate identity preservation using cosine similarity (CSIM)~\cite{doukas2021headgan, gao2023high, siarohin2019first, wang2021one}.
However, we have found these methods to be often unreliable and unpredictable for evaluating motion on a larger scale. CSIM tends to underestimate identity similarity in cases of significant head poses and overestimate it when face images exhibit warping distortions that alter facial shape. Figure \ref{fig:csim} presents some illustrative examples: images of the same individual with large poses show low CSIM scores, while images with noticeable warping artifacts receive high CSIM scores. Our metric for measuring shape error complements the CSIM metric and offers greater robustness against pose variations.

Evaluating motion accuracy in cross-identity motion transfer poses a significant challenge due to the difficulty in obtaining ground truth results for this generative task.
Existing methods have resorted to indirect metrics, with ARD (average rotation distance)\cite{doukas2021headgan, gao2023high} and AUH (average facial action unit hamming distance)\cite{doukas2021headgan, wang2021one} being the most commonly used.
The ARD metric focuses exclusively on head poses, particularly head rotations, whereas the AUH metric, as proposed in \cite{doukas2021headgan}, quantifies the binary hamming distance between two boolean vectors derived from a subset of the facial action coding system (FACS). Our motion error metric offers a comprehensive assessment of head poses, facial expressions, and additional motion nuances.

Finally, we have observed that each metric proposed in previous studies is capable of evaluating only a single aspect of cross-identity motion transfer quality. Our cross-transfer error metric addresses this gap by assessing the overall quality of cross-identity motion transfer.





\begin{table*}[htbp]
\centering
\caption{Hyperparameters of representation learning model on LRS3 dataset.}
\label{tab:hyperparameters_representation}
\begin{tabular}{>{\raggedright}p{8cm} >{\raggedright\arraybackslash}p{4cm}}
\toprule
\textbf{Hyperparameter} & \textbf{Value} \\

\midrule
\multicolumn{2}{c}{\textbf{Disentangle Encoder}} \\Architecture & T5 Encoder \\
Transformer hidden size & 512 \\
Transformer feed-forward dimension & 2048 \\
Transformer layers & 12 \\
Attention heads & 8 \\

\midrule
\multicolumn{2}{c}{\textbf{Denoise Decoder}} \\
Architecture & DiT-B/4 \\
Resolution & 32 \\
Input channels & 4 \\
Patch size & 4 \\
Transformer hidden size & 768 \\
Transformer feed-forward dimension & 3072 \\
Transformer layers & 12 \\
Attention heads of spatial block & 12 \\
Attention heads of temporal block & 12 \\
Temporal block indices & [0, 2, 4, 6, 8, 10] \\
Classifier-free guidance masking rate & 0.1 \\


\midrule
\multicolumn{2}{c}{\textbf{Sampling}} \\
Strategy & EDM \\
Classifier-free guidance & 1.5 \\
Diffusion denoising steps & 128 \\
\midrule
\multicolumn{2}{c}{\textbf{Bitrate-controlled Vector Quantization}} \\
Codebook numbers & 64 \\
Code dimension per codebook & 16 \\
Number of entries per codebook  & 32 \\
Target bitrate & 4kbps \\
Maximum of gumbel-softmax temperature & 1.5 \\
Minimum of gumbel-softmax temperature & 0.1 \\
Temperature decaying rate & 0.9999972 \\

\midrule
\multicolumn{2}{c}{\textbf{Optimization}} \\
Optimizer & AdamW \\
Learning Rate & 2e-4 for denoise decoder, 1e-4 for others \\
Batch size & 32 (4$\times$A100 GPU$\times$8)\\
Weight Decay & 0.01 \\
$\beta$ & (0.9,0.999) \\

\bottomrule
\end{tabular}
\end{table*}

\begin{table*}[htbp]
\centering
\caption{Hyperparameters of representation learning model on Sprites dataset.}
\label{tab:hyperparameters_representation_sprites}
\begin{tabular}{>{\raggedright}p{8cm} >{\raggedright\arraybackslash}p{4cm}}
\toprule
\textbf{Hyperparameter} & \textbf{Value} \\

\midrule
\multicolumn{2}{c}{\textbf{Disentangle Encoder}} \\Architecture & T5 Encoder \\
Transformer hidden size & 384 \\
Transformer feed-forward dimension & 1536 \\
Transformer layers & 12 \\
Attention heads & 6 \\

\midrule
\multicolumn{2}{c}{\textbf{Denoise Decoder}} \\
Architecture & DiT-S/8 \\
Resolution & 64 \\
Input channels & 3 \\
Patch size & 8 \\
Transformer hidden size & 384 \\
Transformer feed-forward dimension & 1536 \\
Transformer layers & 12 \\
Attention heads of spatial block & 6 \\
Attention heads of temporal block & 6 \\
Temporal block indices & [0, 2, 4, 6, 8, 10] \\
Classifier-free guidance masking rate & 0.1 \\

\midrule
\multicolumn{2}{c}{\textbf{Sampling}} \\
Strategy & EDM \\
Classifier-free guidance & 1.5 \\
Diffusion denoising steps & 50 \\
\midrule
\multicolumn{2}{c}{\textbf{Bitrate-controlled Vector Quantization}} \\
Codebook numbers & 1 \\
Code dimension per codebook & 384 \\
Number of entries per codebook  & 64 \\
Target bitrate & 150bps \\
Maximum of gumbel-softmax temperature & 2.0 \\
Minimum of gumbel-softmax temperature & 0.5 \\
Temperature decaying rate & 0.9999972 \\

\midrule
\multicolumn{2}{c}{\textbf{Optimization}} \\
Optimizer & AdamW \\
Learning Rate & 5e-5 \\
Batch size & 128 \\
Weight Decay & 0.01 \\
$\beta$ & (0.9,0.999) \\

\bottomrule
\end{tabular}
\end{table*}

\begin{table*}[htbp]
\centering
\caption{Hyperparameters of motion generation model on LRS3 dataset.}
\label{tab:hyperparameters_gpt}
\begin{tabular}{>{\raggedright}p{8cm} >{\raggedright\arraybackslash}p{4cm}}
\toprule
\textbf{Hyperparameter} & \textbf{Value} \\
\midrule
\multicolumn{2}{c}{\textbf{Model}} \\
Architecture & GPT-2 \\
Hidden size & 1024 \\
Layers & 12 \\
Heads & 8 \\

\midrule
\multicolumn{2}{c}{\textbf{Optimization}} \\
Optimizer & AdamW \\
Learning Rate & 0.0001 \\
Batch size & 256 (8$\times$H100 GPU $\times$ 32) \\
Weight Decay & 0.01 \\
$\beta$ & (0.9,0.999) \\

\bottomrule
\end{tabular}
\end{table*}

\clearpage
\twocolumn

%% file: figures/supp_sprites.tex
\begin{figure}[t]
\centering
\begin{minipage}{0.48\linewidth}
\centering
\tiny Input A
\end{minipage}
\begin{minipage}{0.48\linewidth}
\centering
\tiny Input B
\end{minipage}
\\
\begin{minipage}{0.48\linewidth}
\includegraphics[width=\textwidth]{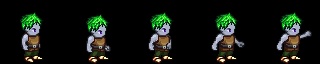}     
\end{minipage}
\begin{minipage}{0.48\linewidth}
\includegraphics[width=\textwidth]{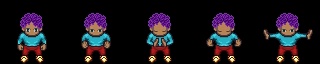} 
\end{minipage}
\\
\begin{minipage}{0.48\linewidth}
\centering
\tiny motion of A + content of B
\end{minipage}
\begin{minipage}{0.48\linewidth}
\centering
\tiny motion of B + content of A
\end{minipage}
\\
\begin{minipage}{0.48\linewidth}
\includegraphics[width=\textwidth]{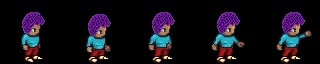} 
\end{minipage}
\begin{minipage}{0.48\linewidth}
\includegraphics[width=\textwidth]{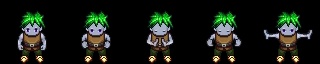} 
\end{minipage}
\\

\vspace{2pt}
\begin{minipage}{0.48\linewidth}
\centering
\tiny Input A
\end{minipage}
\begin{minipage}{0.48\linewidth}
\centering
\tiny Input B
\end{minipage}
\\
\begin{minipage}{0.48\linewidth}
\includegraphics[width=\textwidth]{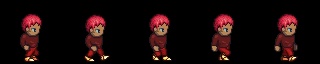}     
\end{minipage}
\begin{minipage}{0.48\linewidth}
\includegraphics[width=\textwidth]{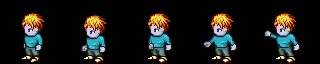} 
\end{minipage}
\\
\begin{minipage}{0.48\linewidth}
\centering
\tiny motion of A + content of B
\end{minipage}
\begin{minipage}{0.48\linewidth}
\centering
\tiny motion of B + content of A
\end{minipage}
\\
\begin{minipage}{0.48\linewidth}
\includegraphics[width=\textwidth]{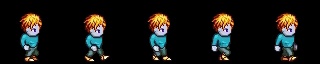} 
\end{minipage}
\begin{minipage}{0.48\linewidth}
\includegraphics[width=\textwidth]{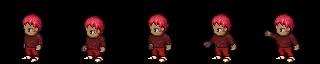} 
\end{minipage}
\\

\vspace{2pt}
\begin{minipage}{0.48\linewidth}
\centering
\tiny Input A
\end{minipage}
\begin{minipage}{0.48\linewidth}
\centering
\tiny Input B
\end{minipage}
\\
\begin{minipage}{0.48\linewidth}
\includegraphics[width=\textwidth]{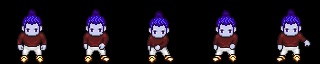}     
\end{minipage}
\begin{minipage}{0.48\linewidth}
\includegraphics[width=\textwidth]{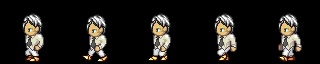} 
\end{minipage}
\\
\begin{minipage}{0.48\linewidth}
\centering
\tiny motion of A + content of B
\end{minipage}
\begin{minipage}{0.48\linewidth}
\centering
\tiny motion of B + content of A
\end{minipage}
\\
\begin{minipage}{0.48\linewidth}
\includegraphics[width=\textwidth]{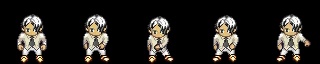} 
\end{minipage}
\begin{minipage}{0.48\linewidth}
\includegraphics[width=\textwidth]{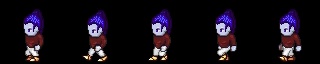} 
\end{minipage}
\\
\caption{Additional motion transfer results on Sprites test set.}
\label{fig:supp_sprites}
\end{figure}

%% file: tables/results_sprites.tex
\begin{table}[htb]
\centering
\caption{Cross-driven attributes accuracies(\%) of different methods on Sprites test set. ``Action" represents the motion attribute and the others are content attributes.}
\label{tab:results_sprites}
\scalebox{0.85}{ 
\begin{tabular}{@{}cccccc@{}}
\toprule
Methods & Action$\uparrow$ & Skin$\uparrow$ & Pants$\uparrow$ & Top$\uparrow$ & Hair$\uparrow$ \\
\midrule
C-DSVAE & 100.00 & 100.00 & 100.00 & 100.00 & 100.00\\
Ours & 100.00 & 100.00 & 100.00 & 100.00 & 100.00\\

 \bottomrule
\end{tabular}
}
\end{table}

%% file: figures/ablation_bitrate.tex
\begin{figure*}[bth]
\centering
\begin{minipage}{0.13\linewidth}
\centering
\scriptsize Content Ref.
\end{minipage}
\begin{minipage}{0.13\linewidth}
\centering
\scriptsize Motion Ref.
\end{minipage}
\begin{minipage}{0.13\linewidth}
\centering
\scriptsize 2kbps
\end{minipage}
\begin{minipage}{0.13\linewidth}
\centering
\scriptsize 6kbps
\end{minipage}
\begin{minipage}{0.13\linewidth}
\centering
\scriptsize 8kbps
\end{minipage}
\begin{minipage}{0.13\linewidth}
\centering
\scriptsize 16kbps
\end{minipage}
\begin{minipage}{0.13\linewidth}
\centering
\scriptsize \textbf{4kbps}
\end{minipage}
\\
\begin{minipage}{0.13\linewidth}
\includegraphics[width=\textwidth]{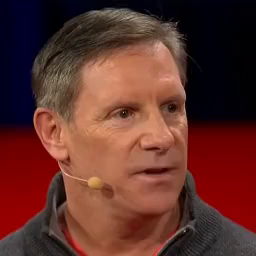} 
\end{minipage}
\begin{minipage}{0.13\linewidth}
\includegraphics[width=\textwidth]{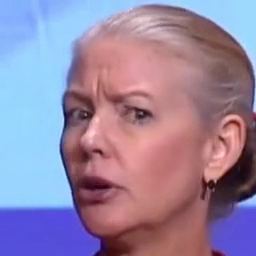} 
\end{minipage}
\begin{minipage}{0.13\linewidth}
\includegraphics[width=\textwidth]{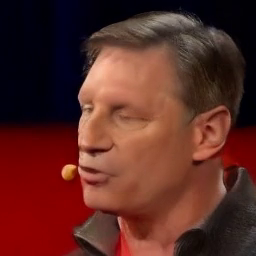} 
\end{minipage}
\begin{minipage}{0.13\linewidth}
\includegraphics[width=\textwidth]{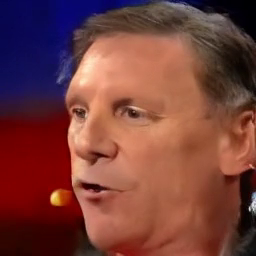} 
\end{minipage}
\begin{minipage}{0.13\linewidth}
\includegraphics[width=\textwidth]{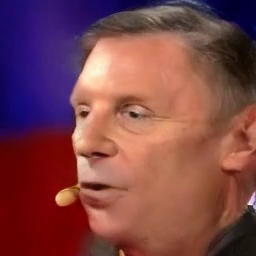} 
\end{minipage}
\begin{minipage}{0.13\linewidth}
\includegraphics[width=\textwidth]{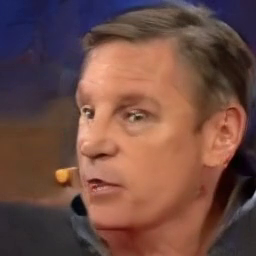} 
\end{minipage}
\begin{minipage}{0.13\linewidth}
\includegraphics[width=\textwidth]{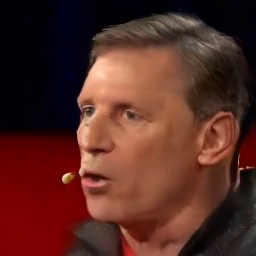} 
\end{minipage}
\\
\begin{minipage}{0.13\linewidth}
\includegraphics[width=\textwidth]{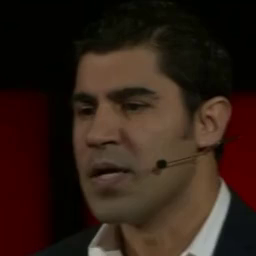} 
\end{minipage}
\begin{minipage}{0.13\linewidth}
\includegraphics[width=\textwidth]{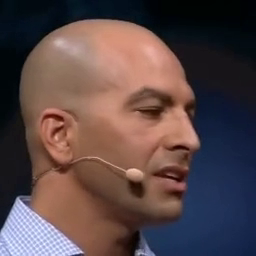} 
\end{minipage}
\begin{minipage}{0.13\linewidth}
\includegraphics[width=\textwidth]{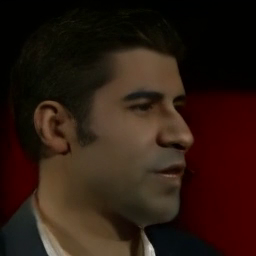} 
\end{minipage}
\begin{minipage}{0.13\linewidth}
\includegraphics[width=\textwidth]{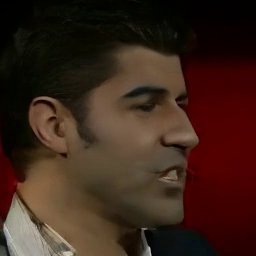} 
\end{minipage}
\begin{minipage}{0.13\linewidth}
\includegraphics[width=\textwidth]{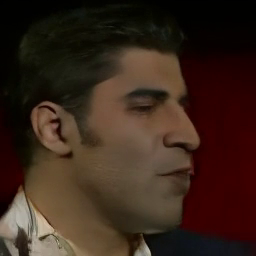} 
\end{minipage}
\begin{minipage}{0.13\linewidth}
\includegraphics[width=\textwidth]{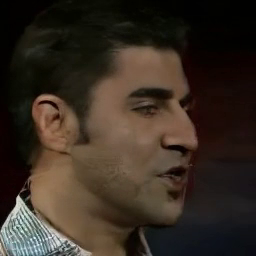} 
\end{minipage}
\begin{minipage}{0.13\linewidth}
\includegraphics[width=\textwidth]{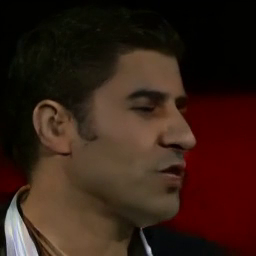} 
\end{minipage}
\\
\begin{minipage}{0.13\linewidth}
\includegraphics[width=\textwidth]{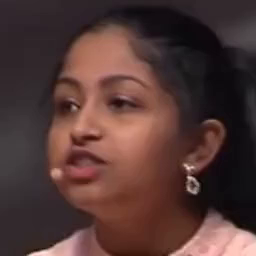} 
\end{minipage}
\begin{minipage}{0.13\linewidth}
\includegraphics[width=\textwidth]{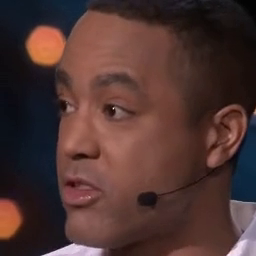} 
\end{minipage}
\begin{minipage}{0.13\linewidth}
\includegraphics[width=\textwidth]{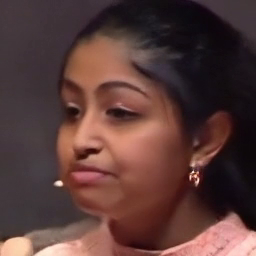} 
\end{minipage}
\begin{minipage}{0.13\linewidth}
\includegraphics[width=\textwidth]{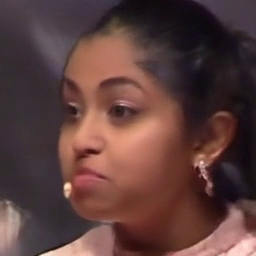} 
\end{minipage}
\begin{minipage}{0.13\linewidth}
\includegraphics[width=\textwidth]{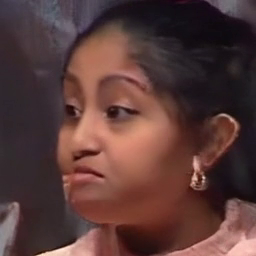} 
\end{minipage}
\begin{minipage}{0.13\linewidth}
\includegraphics[width=\textwidth]{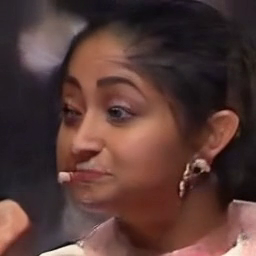} 
\end{minipage}
\begin{minipage}{0.13\linewidth}
\includegraphics[width=\textwidth]{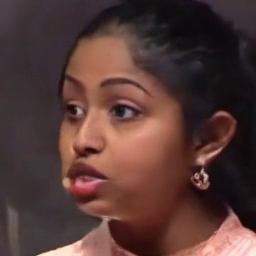} 
\end{minipage}
\\
\begin{minipage}{0.13\linewidth}
\includegraphics[width=\textwidth]{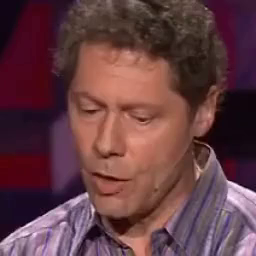} 
\end{minipage}
\begin{minipage}{0.13\linewidth}
\includegraphics[width=\textwidth]{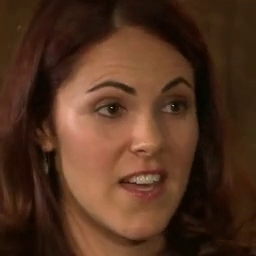} 
\end{minipage}
\begin{minipage}{0.13\linewidth}
\includegraphics[width=\textwidth]{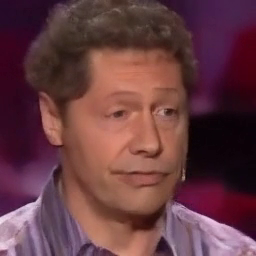} 
\end{minipage}
\begin{minipage}{0.13\linewidth}
\includegraphics[width=\textwidth]{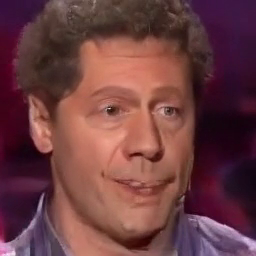} 
\end{minipage}
\begin{minipage}{0.13\linewidth}
\includegraphics[width=\textwidth]{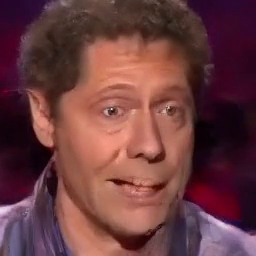} 
\end{minipage}
\begin{minipage}{0.13\linewidth}
\includegraphics[width=\textwidth]{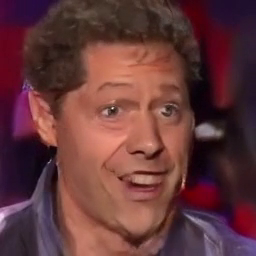} 
\end{minipage}
\begin{minipage}{0.13\linewidth}
\includegraphics[width=\textwidth]{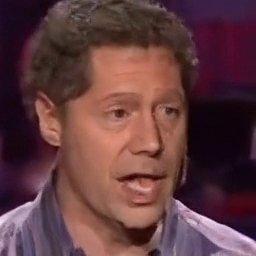} 
\end{minipage}
\\

\caption{Ablation study on target bitrate selection. 
\textit{Column 1-2}: Content and motion reference.
\textit{Column 3-7}: Motion transfer results under different target bitrate. Bitrate 4kbps achieves the best tradeoff between image fidelity and disentanglement.}
\label{fig:ablation_bitrate}
\end{figure*}

%% file: tables/ablation_content.tex

\begin{table}[htb]
\centering
\caption{Quantitative comparison for motion transfer with different content feature frames.}
\label{tab:ablation_content}
\scalebox{0.7}{ 
\begin{tabular}{@{}cccccc@{}}
\toprule
Frames             & FID$\downarrow$ & CSIM$\uparrow$ & \begin{tabular}[c]{@{}c@{}}Shape error$\downarrow$\\ $\times 10^{-1}$\end{tabular} & \begin{tabular}[c]{@{}c@{}}Motion error$\downarrow$\\ $\times 10^{-2}$\end{tabular} & \begin{tabular}[c]{@{}c@{}}Cross error$\downarrow$\\ $\times 10^{-2}$\end{tabular} \\ \midrule
1         & 87.9 & \textbf{0.692} & 0.47 & 3.13 & 3.81 \\
5         & 86.0 & 0.685 & 0.43 & 3.50 & 4.21 \\
15        & \textbf{85.6} & 0.685 & 0.43 & 3.47 & 4.14 \\
20        & 86.6 & 0.688 & 0.44 & 3.20 & 3.78 \\
\midrule
all       & 86.0 & \textbf{0.692} & \textbf{0.41} & \textbf{3.13} & \textbf{3.67} \\
\bottomrule
\end{tabular}
}
\end{table}

%% file: tables/ablation_rebuttal.tex
\begin{table}[htb]
\centering
\caption{Ablation study of the usage of bitrate loss.}
\label{tab:ablation_rebuttal}
\scalebox{0.6}{ 
\begin{tabular}{@{}cccccc@{}}
\toprule
Method             & FID$\downarrow$ & CSIM$\uparrow$ & \begin{tabular}[c]{@{}c@{}}Shape error$\downarrow$\\ $\times 10^{-1}$\end{tabular} & \begin{tabular}[c]{@{}c@{}}Motion error$\downarrow$\\ $\times 10^{-2}$\end{tabular} & \begin{tabular}[c]{@{}c@{}}Cross error$\downarrow$\\ $\times 10^{-2}$\end{tabular} \\ \midrule
w/o bitrate loss  & 92.7 & \textbf{0.70} & 0.59  & 4.81 & 6.21      \\
\midrule
Ours  & \textbf{86.0} & 0.69 & \textbf{0.41} & \textbf{3.13} & \textbf{3.67} \\
\bottomrule
\end{tabular}
}
\end{table}

%% file: figures/csim.tex
\begin{figure}[bth]
\centering
\begin{minipage}{0.32\linewidth}
\includegraphics[width=\textwidth]{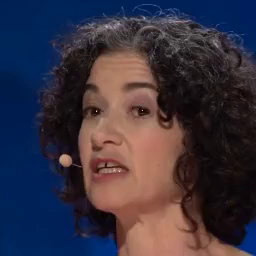} 
\end{minipage}
\begin{minipage}{0.32\linewidth}
\includegraphics[width=\textwidth]{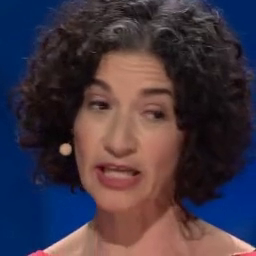} 
\end{minipage}
\begin{minipage}{0.32\linewidth}
\includegraphics[width=\textwidth]{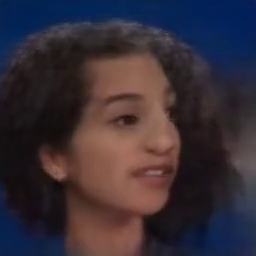} 
\end{minipage}
\\
\vspace{0.1cm}
\begin{minipage}{0.32\linewidth}
\centering
\scriptsize Reference CSIM=1.000
\end{minipage}
\begin{minipage}{0.32\linewidth}
\centering
\scriptsize CSIM=0.299
\end{minipage}
\begin{minipage}{0.32\linewidth}
\centering
\scriptsize CSIM=0.501
\end{minipage}
\\
\vspace{0.35cm}
\begin{minipage}{0.32\linewidth}
\includegraphics[width=\textwidth]{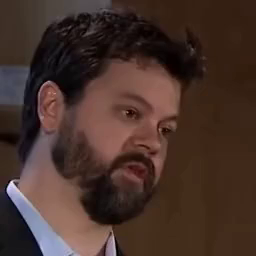} 
\end{minipage}
\begin{minipage}{0.32\linewidth}
\includegraphics[width=\textwidth]{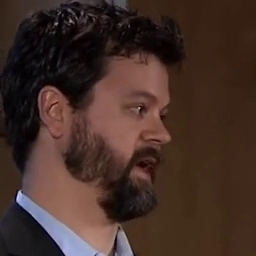} 
\end{minipage}
\begin{minipage}{0.32\linewidth}
\includegraphics[width=\textwidth]{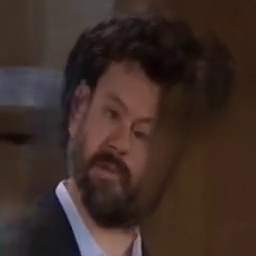} 
\end{minipage}
\\
\vspace{0.1cm}
\begin{minipage}{0.32\linewidth}
\centering
\scriptsize Reference CSIM=1.000
\end{minipage}
\begin{minipage}{0.32\linewidth}
\centering
\scriptsize CSIM=0.211
\end{minipage}
\begin{minipage}{0.32\linewidth}
\centering
\scriptsize CSIM=0.559
\end{minipage}
\\

\vspace{0.35cm}
\begin{minipage}{0.32\linewidth}
\includegraphics[width=\textwidth]{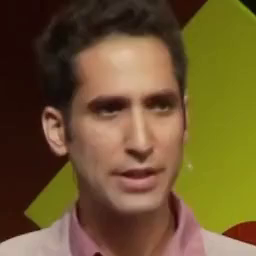} 
\end{minipage}
\begin{minipage}{0.32\linewidth}
\includegraphics[width=\textwidth]{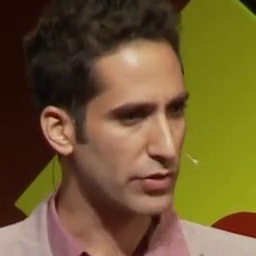} 
\end{minipage}
\begin{minipage}{0.32\linewidth}
\includegraphics[width=\textwidth]{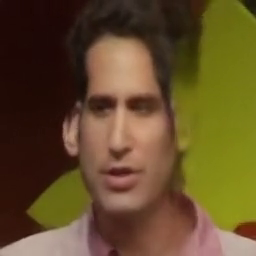} 
\end{minipage}
\\
\vspace{0.1cm}
\begin{minipage}{0.32\linewidth}
\centering
\scriptsize Reference CSIM=1.000
\end{minipage}
\begin{minipage}{0.32\linewidth}
\centering
\scriptsize CSIM=0.210
\end{minipage}
\begin{minipage}{0.32\linewidth}
\centering
\scriptsize CSIM=0.504
\end{minipage}
\\

\caption{CSIM analysis. Images with the same identities but different poses compared to the reference images get low CSIM, but images with obvious warping artifacts achieve high CSIM, indicating the probably unreliable evaluation results of the CSIM metric.}
\label{fig:csim}
\end{figure}

%% file: main.bbl
\begin{thebibliography}{66}
\providecommand{\natexlab}[1]{#1}
\providecommand{\url}[1]{\texttt{#1}}
\expandafter\ifx\csname urlstyle\endcsname\relax
  \providecommand{\doi}[1]{doi: #1}\else
  \providecommand{\doi}{doi: \begingroup \urlstyle{rm}\Url}\fi

\bibitem[Achille and Soatto(2018)]{achille2018emergence}
Alessandro Achille and Stefano Soatto.
\newblock Emergence of invariance and disentanglement in deep representations.
\newblock \emph{The Journal of Machine Learning Research}, 19\penalty0 (1):\penalty0 1947--1980, 2018.

\bibitem[Afouras et~al.(2018)Afouras, Chung, and Zisserman]{afouras2018lrs3}
Triantafyllos Afouras, Joon~Son Chung, and Andrew Zisserman.
\newblock Lrs3-ted: a large-scale dataset for visual speech recognition.
\newblock \emph{arXiv preprint arXiv:1809.00496}, 2018.

\bibitem[Alemi et~al.(2016)Alemi, Fischer, Dillon, and Murphy]{alemi2016deep}
Alexander~A Alemi, Ian Fischer, Joshua~V Dillon, and Kevin Murphy.
\newblock Deep variational information bottleneck.
\newblock \emph{arXiv preprint arXiv:1612.00410}, 2016.

\bibitem[Baevski et~al.(2019)Baevski, Schneider, and Auli]{baevski2019vq}
Alexei Baevski, Steffen Schneider, and Michael Auli.
\newblock vq-wav2vec: Self-supervised learning of discrete speech representations.
\newblock \emph{arXiv preprint arXiv:1910.05453}, 2019.

\bibitem[Bai et~al.(2021)Bai, Wang, and Gomes]{bai2021contrastively}
Junwen Bai, Weiran Wang, and Carla~P Gomes.
\newblock Contrastively disentangled sequential variational autoencoder.
\newblock \emph{Advances in Neural Information Processing Systems}, 34:\penalty0 10105--10118, 2021.

\bibitem[Berman et~al.(2023)Berman, Naiman, and Azencot]{berman2023multifactor}
Nimrod Berman, Ilan Naiman, and Omri Azencot.
\newblock Multifactor sequential disentanglement via structured koopman autoencoders.
\newblock \emph{arXiv preprint arXiv:2303.17264}, 2023.

\bibitem[Berman et~al.(2024)Berman, Naiman, Arbiv, Fadlon, and Azencot]{berman2024sequential}
Nimrod Berman, Ilan Naiman, Idan Arbiv, Gal Fadlon, and Omri Azencot.
\newblock Sequential disentanglement by extracting static information from a single sequence element.
\newblock \emph{arXiv preprint arXiv:2406.18131}, 2024.

\bibitem[Blattmann et~al.(2023)Blattmann, Rombach, Ling, Dockhorn, Kim, Fidler, and Kreis]{blattmann2023align}
Andreas Blattmann, Robin Rombach, Huan Ling, Tim Dockhorn, Seung~Wook Kim, Sanja Fidler, and Karsten Kreis.
\newblock Align your latents: High-resolution video synthesis with latent diffusion models.
\newblock In \emph{Proceedings of the IEEE/CVF Conference on Computer Vision and Pattern Recognition}, pages 22563--22575, 2023.

\bibitem[Bounareli et~al.(2023)Bounareli, Tzelepis, Argyriou, Patras, and Tzimiropoulos]{bounareli2023hyperreenact}
Stella Bounareli, Christos Tzelepis, Vasileios Argyriou, Ioannis Patras, and Georgios Tzimiropoulos.
\newblock Hyperreenact: One-shot reenactment via jointly learning to refine and retarget faces.
\newblock In \emph{Proceedings of the IEEE/CVF International Conference on Computer Vision (ICCV)}, 2023.

\bibitem[Brooks et~al.(2024)Brooks, Peebles, Holmes, DePue, Guo, Jing, Schnurr, Taylor, Luhman, Luhman, Ng, Wang, and Ramesh]{videoworldsimulators2024}
Tim Brooks, Bill Peebles, Connor Holmes, Will DePue, Yufei Guo, Li Jing, David Schnurr, Joe Taylor, Troy Luhman, Eric Luhman, Clarence Ng, Ricky Wang, and Aditya Ramesh.
\newblock Video generation models as world simulators.
\newblock 2024.

\bibitem[Chen et~al.(2020)Chen, Kornblith, Norouzi, and Hinton]{chen2020simple}
Ting Chen, Simon Kornblith, Mohammad Norouzi, and Geoffrey Hinton.
\newblock A simple framework for contrastive learning of visual representations.
\newblock In \emph{International conference on machine learning}, pages 1597--1607. PMLR, 2020.

\bibitem[Cover(1999)]{cover1999elements}
Thomas~M Cover.
\newblock \emph{Elements of information theory}.
\newblock John Wiley \& Sons, 1999.

\bibitem[Deng et~al.(2019{\natexlab{a}})Deng, Guo, Niannan, and Zafeiriou]{deng2018arcface}
Jiankang Deng, Jia Guo, Xue Niannan, and Stefanos Zafeiriou.
\newblock Arcface: Additive angular margin loss for deep face recognition.
\newblock In \emph{CVPR}, 2019{\natexlab{a}}.

\bibitem[Deng et~al.(2019{\natexlab{b}})Deng, Yang, Xu, Chen, Jia, and Tong]{deng2019accurate}
Yu Deng, Jiaolong Yang, Sicheng Xu, Dong Chen, Yunde Jia, and Xin Tong.
\newblock Accurate 3d face reconstruction with weakly-supervised learning: From single image to image set.
\newblock In \emph{Proceedings of the IEEE/CVF conference on computer vision and pattern recognition workshops}, pages 0--0, 2019{\natexlab{b}}.

\bibitem[Dosovitskiy et~al.(2020)Dosovitskiy, Beyer, Kolesnikov, Weissenborn, Zhai, Unterthiner, Dehghani, Minderer, Heigold, Gelly, et~al.]{dosovitskiy2020image}
Alexey Dosovitskiy, Lucas Beyer, Alexander Kolesnikov, Dirk Weissenborn, Xiaohua Zhai, Thomas Unterthiner, Mostafa Dehghani, Matthias Minderer, Georg Heigold, Sylvain Gelly, et~al.
\newblock An image is worth 16x16 words: Transformers for image recognition at scale.
\newblock \emph{arXiv preprint arXiv:2010.11929}, 2020.

\bibitem[Doukas et~al.(2021)Doukas, Zafeiriou, and Sharmanska]{doukas2021headgan}
Michail~Christos Doukas, Stefanos Zafeiriou, and Viktoriia Sharmanska.
\newblock Headgan: One-shot neural head synthesis and editing.
\newblock In \emph{Proceedings of the IEEE/CVF International conference on Computer Vision}, pages 14398--14407, 2021.

\bibitem[Esser et~al.(2021)Esser, Rombach, and Ommer]{esser2021taming}
Patrick Esser, Robin Rombach, and Bjorn Ommer.
\newblock Taming transformers for high-resolution image synthesis.
\newblock In \emph{Proceedings of the IEEE/CVF conference on computer vision and pattern recognition}, pages 12873--12883, 2021.

\bibitem[Gao et~al.(2023)Gao, Zhou, Wang, Li, Ming, and Lu]{gao2023high}
Yue Gao, Yuan Zhou, Jinglu Wang, Xiao Li, Xiang Ming, and Yan Lu.
\newblock High-fidelity and freely controllable talking head video generation.
\newblock In \emph{Proceedings of the IEEE/CVF Conference on Computer Vision and Pattern Recognition}, pages 5609--5619, 2023.

\bibitem[Gao et~al.(2024)Gao, Li, Chu, and Lu]{gao2024implicit}
Yue Gao, Jiahao Li, Lei Chu, and Yan Lu.
\newblock Implicit motion function.
\newblock In \emph{Proceedings of the IEEE/CVF Conference on Computer Vision and Pattern Recognition}, pages 19278--19289, 2024.

\bibitem[Goodfellow et~al.(2014)Goodfellow, Pouget-Abadie, Mirza, Xu, Warde-Farley, Ozair, Courville, and Bengio]{goodfellow2014generative}
Ian Goodfellow, Jean Pouget-Abadie, Mehdi Mirza, Bing Xu, David Warde-Farley, Sherjil Ozair, Aaron Courville, and Yoshua Bengio.
\newblock Generative adversarial nets.
\newblock \emph{Advances in neural information processing systems}, 27, 2014.

\bibitem[Grassal et~al.(2022)Grassal, Prinzler, Leistner, Rother, Nie{\ss}ner, and Thies]{grassal2022neural}
Philip-William Grassal, Malte Prinzler, Titus Leistner, Carsten Rother, Matthias Nie{\ss}ner, and Justus Thies.
\newblock Neural head avatars from monocular rgb videos.
\newblock In \emph{Proceedings of the IEEE/CVF conference on computer vision and pattern recognition}, pages 18653--18664, 2022.

\bibitem[Grill et~al.(2020)Grill, Strub, Altch{\'e}, Tallec, Richemond, Buchatskaya, Doersch, Avila~Pires, Guo, Gheshlaghi~Azar, et~al.]{grill2020bootstrap}
Jean-Bastien Grill, Florian Strub, Florent Altch{\'e}, Corentin Tallec, Pierre Richemond, Elena Buchatskaya, Carl Doersch, Bernardo Avila~Pires, Zhaohan Guo, Mohammad Gheshlaghi~Azar, et~al.
\newblock Bootstrap your own latent-a new approach to self-supervised learning.
\newblock \emph{Advances in neural information processing systems}, 33:\penalty0 21271--21284, 2020.

\bibitem[Gu et~al.(2022)Gu, Chen, Bao, Wen, Zhang, Chen, Yuan, and Guo]{gu2022vector}
Shuyang Gu, Dong Chen, Jianmin Bao, Fang Wen, Bo Zhang, Dongdong Chen, Lu Yuan, and Baining Guo.
\newblock Vector quantized diffusion model for text-to-image synthesis.
\newblock In \emph{Proceedings of the IEEE/CVF Conference on Computer Vision and Pattern Recognition}, pages 10696--10706, 2022.

\bibitem[Guo et~al.(2024)Guo, Zhang, Liu, Zhong, Zhang, Wan, and Zhang]{guo2024liveportrait}
Jianzhu Guo, Dingyun Zhang, Xiaoqiang Liu, Zhizhou Zhong, Yuan Zhang, Pengfei Wan, and Di Zhang.
\newblock Liveportrait: Efficient portrait animation with stitching and retargeting control.
\newblock \emph{arXiv preprint arXiv:2407.03168}, 2024.

\bibitem[Han et~al.(2021)Han, Min, Han, Li, and Zhang]{han2021disentangled}
Jun Han, Martin~Renqiang Min, Ligong Han, Li~Erran Li, and Xuan Zhang.
\newblock Disentangled recurrent wasserstein autoencoder.
\newblock \emph{arXiv preprint arXiv:2101.07496}, 2021.

\bibitem[He et~al.(2020)He, Fan, Wu, Xie, and Girshick]{he2020momentum}
Kaiming He, Haoqi Fan, Yuxin Wu, Saining Xie, and Ross Girshick.
\newblock Momentum contrast for unsupervised visual representation learning.
\newblock In \emph{Proceedings of the IEEE/CVF conference on computer vision and pattern recognition}, pages 9729--9738, 2020.

\bibitem[Heusel et~al.(2017)Heusel, Ramsauer, Unterthiner, Nessler, and Hochreiter]{heusel2017gans}
Martin Heusel, Hubert Ramsauer, Thomas Unterthiner, Bernhard Nessler, and Sepp Hochreiter.
\newblock Gans trained by a two time-scale update rule converge to a local nash equilibrium.
\newblock \emph{Advances in neural information processing systems}, 30, 2017.

\bibitem[Hewitt et~al.(2024)Hewitt, Saleh, Aliakbarian, Petikam, Rezaeifar, Florentin, Hosenie, Cashman, Valentin, Cosker, and Baltru\v{s}aitis]{hewitt2024look}
Charlie Hewitt, Fatemeh Saleh, Sadegh Aliakbarian, Lohit Petikam, Shideh Rezaeifar, Louis Florentin, Zafiirah Hosenie, Thomas~J Cashman, Julien Valentin, Darren Cosker, and Tadas Baltru\v{s}aitis.
\newblock Look ma, no markers: holistic performance capture without the hassle.
\newblock \emph{ACM Transactions on Graphics (TOG)}, 43\penalty0 (6), 2024.

\bibitem[Ho et~al.(2020)Ho, Jain, and Abbeel]{ho2020denoising}
Jonathan Ho, Ajay Jain, and Pieter Abbeel.
\newblock Denoising diffusion probabilistic models.
\newblock \emph{Advances in neural information processing systems}, 33:\penalty0 6840--6851, 2020.

\bibitem[Hong and Xu(2023)]{hong23implicit}
Fa-Ting Hong and Dan Xu.
\newblock Implicit identity representation conditioned memory compensation network for talking head video generation.
\newblock In \emph{ICCV}, 2023.

\bibitem[Huang et~al.(2022)Huang, Li, Li, Liu, and Lu]{huang2022neural}
Cong Huang, Jiahao Li, Bin Li, Dong Liu, and Yan Lu.
\newblock Neural compression-based feature learning for video restoration.
\newblock In \emph{Proceedings of the IEEE/CVF Conference on Computer Vision and Pattern Recognition}, pages 5872--5881, 2022.

\bibitem[Hudson et~al.(2023)Hudson, Zoran, Malinowski, Lampinen, Jaegle, McClelland, Matthey, Hill, and Lerchner]{hudson2023soda}
Drew~A Hudson, Daniel Zoran, Mateusz Malinowski, Andrew~K Lampinen, Andrew Jaegle, James~L McClelland, Loic Matthey, Felix Hill, and Alexander Lerchner.
\newblock Soda: Bottleneck diffusion models for representation learning.
\newblock \emph{arXiv preprint arXiv:2311.17901}, 2023.

\bibitem[Jiang et~al.(2023)Jiang, Peng, Zhang, and Lu]{jiang2023disentangled}
Xue Jiang, Xiulian Peng, Yuan Zhang, and Yan Lu.
\newblock Disentangled feature learning for real-time neural speech coding.
\newblock In \emph{ICASSP 2023-2023 IEEE International Conference on Acoustics, Speech and Signal Processing (ICASSP)}, pages 1--5. IEEE, 2023.

\bibitem[Karras et~al.(2022)Karras, Aittala, Aila, and Laine]{karras2022elucidating}
Tero Karras, Miika Aittala, Timo Aila, and Samuli Laine.
\newblock Elucidating the design space of diffusion-based generative models.
\newblock \emph{Advances in Neural Information Processing Systems}, 35:\penalty0 26565--26577, 2022.

\bibitem[Li et~al.(2021{\natexlab{a}})Li, Li, and Lu]{li2021deep}
Jiahao Li, Bin Li, and Yan Lu.
\newblock Deep contextual video compression.
\newblock \emph{Advances in Neural Information Processing Systems}, 34:\penalty0 18114--18125, 2021{\natexlab{a}}.

\bibitem[Li et~al.(2022)Li, Li, and Lu]{li2022hybrid}
Jiahao Li, Bin Li, and Yan Lu.
\newblock Hybrid spatial-temporal entropy modelling for neural video compression.
\newblock In \emph{Proceedings of the 30th ACM International Conference on Multimedia}, 2022.

\bibitem[Li et~al.(2023{\natexlab{a}})Li, Li, and Lu]{li2023neural}
Jiahao Li, Bin Li, and Yan Lu.
\newblock Neural video compression with diverse contexts.
\newblock In \emph{Proceedings of the IEEE/CVF Conference on Computer Vision and Pattern Recognition}, pages 22616--22626, 2023{\natexlab{a}}.

\bibitem[Li et~al.(2023{\natexlab{b}})Li, Li, Savarese, and Hoi]{li2023blip}
Junnan Li, Dongxu Li, Silvio Savarese, and Steven Hoi.
\newblock Blip-2: Bootstrapping language-image pre-training with frozen image encoders and large language models.
\newblock \emph{arXiv preprint arXiv:2301.12597}, 2023{\natexlab{b}}.

\bibitem[Li et~al.(2021{\natexlab{b}})Li, Zhang, Qiu, Yao, Liu, and Mei]{li2021motion}
Rui Li, Yiheng Zhang, Zhaofan Qiu, Ting Yao, Dong Liu, and Tao Mei.
\newblock Motion-focused contrastive learning of video representations.
\newblock In \emph{Proceedings of the IEEE/CVF International Conference on Computer Vision}, pages 2105--2114, 2021{\natexlab{b}}.

\bibitem[Li et~al.(2023{\natexlab{c}})Li, Katabi, and He]{li2023self}
Tianhong Li, Dina Katabi, and Kaiming He.
\newblock Self-conditioned image generation via generating representations.
\newblock \emph{arXiv preprint arXiv:2312.03701}, 2023{\natexlab{c}}.

\bibitem[Li and Mandt(2018)]{li2018disentangled}
Yingzhen Li and Stephan Mandt.
\newblock Disentangled sequential autoencoder.
\newblock \emph{arXiv preprint arXiv:1803.02991}, 2018.

\bibitem[Liu et~al.(2024)Liu, Chen, Fan, Du, Chen, Chen, and Yu]{liu2024anitalker}
Tao Liu, Feilong Chen, Shuai Fan, Chenpeng Du, Qi Chen, Xie Chen, and Kai Yu.
\newblock Anitalker: animate vivid and diverse talking faces through identity-decoupled facial motion encoding.
\newblock In \emph{Proceedings of the 32nd ACM International Conference on Multimedia}, pages 6696--6705, 2024.

\bibitem[Loshchilov and Hutter(2017)]{loshchilov2017decoupled}
Ilya Loshchilov and Frank Hutter.
\newblock Decoupled weight decay regularization.
\newblock \emph{arXiv preprint arXiv:1711.05101}, 2017.

\bibitem[Mallya et~al.(2022)Mallya, Wang, and Liu]{mallya2022implicit}
Arun Mallya, Ting-Chun Wang, and Ming-Yu Liu.
\newblock Implicit warping for animation with image sets.
\newblock \emph{Advances in Neural Information Processing Systems}, 35:\penalty0 22438--22450, 2022.

\bibitem[Naiman et~al.(2023)Naiman, Berman, and Azencot]{naiman2023sample}
Ilan Naiman, Nimrod Berman, and Omri Azencot.
\newblock Sample and predict your latent: modality-free sequential disentanglement via contrastive estimation.
\newblock In \emph{International Conference on Machine Learning}, pages 25694--25717. PMLR, 2023.

\bibitem[Peebles and Xie(2023)]{peebles2023scalable}
William Peebles and Saining Xie.
\newblock Scalable diffusion models with transformers.
\newblock In \emph{Proceedings of the IEEE/CVF International Conference on Computer Vision}, pages 4195--4205, 2023.

\bibitem[Peng et~al.(2022)Peng, Dong, Bao, Ye, and Wei]{peng2022beit}
Zhiliang Peng, Li Dong, Hangbo Bao, Qixiang Ye, and Furu Wei.
\newblock Beit v2: Masked image modeling with vector-quantized visual tokenizers.
\newblock \emph{arXiv preprint arXiv:2208.06366}, 2022.

\bibitem[Radford et~al.(2019)Radford, Wu, Child, Luan, Amodei, Sutskever, et~al.]{radford2019language}
Alec Radford, Jeffrey Wu, Rewon Child, David Luan, Dario Amodei, Ilya Sutskever, et~al.
\newblock Language models are unsupervised multitask learners.
\newblock \emph{OpenAI blog}, 1\penalty0 (8):\penalty0 9, 2019.

\bibitem[Raffel et~al.(2020)Raffel, Shazeer, Roberts, Lee, Narang, Matena, Zhou, Li, and Liu]{2020t5}
Colin Raffel, Noam Shazeer, Adam Roberts, Katherine Lee, Sharan Narang, Michael Matena, Yanqi Zhou, Wei Li, and Peter~J. Liu.
\newblock Exploring the limits of transfer learning with a unified text-to-text transformer.
\newblock \emph{Journal of Machine Learning Research}, 21\penalty0 (140):\penalty0 1--67, 2020.

\bibitem[Reed et~al.(2015)Reed, Zhang, Zhang, and Lee]{reed2015deep}
Scott~E Reed, Yi Zhang, Yuting Zhang, and Honglak Lee.
\newblock Deep visual analogy-making.
\newblock \emph{Advances in neural information processing systems}, 28, 2015.

\bibitem[Rombach et~al.(2022)Rombach, Blattmann, Lorenz, Esser, and Ommer]{rombach2022high}
Robin Rombach, Andreas Blattmann, Dominik Lorenz, Patrick Esser, and Bj{\"o}rn Ommer.
\newblock High-resolution image synthesis with latent diffusion models.
\newblock In \emph{Proceedings of the IEEE/CVF conference on computer vision and pattern recognition}, pages 10684--10695, 2022.

\bibitem[Sheng et~al.(2022)Sheng, Li, Li, Li, Liu, and Lu]{sheng2022temporal}
Xihua Sheng, Jiahao Li, Bin Li, Li Li, Dong Liu, and Yan Lu.
\newblock Temporal context mining for learned video compression.
\newblock \emph{IEEE Transactions on Multimedia}, 2022.

\bibitem[Siarohin et~al.(2019)Siarohin, Lathuili{\`e}re, Tulyakov, Ricci, and Sebe]{siarohin2019first}
Aliaksandr Siarohin, St{\'e}phane Lathuili{\`e}re, Sergey Tulyakov, Elisa Ricci, and Nicu Sebe.
\newblock First order motion model for image animation.
\newblock \emph{Advances in neural information processing systems}, 32, 2019.

\bibitem[Simon et~al.(2024)Simon, Frossard, and Vleeschouwer]{simon2024sequential}
Mathieu~Cyrille Simon, Pascal Frossard, and Christophe~De Vleeschouwer.
\newblock Sequential representation learning via static-dynamic conditional disentanglement.
\newblock In \emph{European Conference on Computer Vision}, pages 110--126. Springer, 2024.

\bibitem[Tishby et~al.(2000)Tishby, Pereira, and Bialek]{tishby2000information}
Naftali Tishby, Fernando~C Pereira, and William Bialek.
\newblock The information bottleneck method.
\newblock \emph{arXiv preprint physics/0004057}, 2000.

\bibitem[Van Den~Oord et~al.(2017)Van Den~Oord, Vinyals, et~al.]{van2017neural}
Aaron Van Den~Oord, Oriol Vinyals, et~al.
\newblock Neural discrete representation learning.
\newblock \emph{Advances in neural information processing systems}, 30, 2017.

\bibitem[Vaswani et~al.(2017)Vaswani, Shazeer, Parmar, Uszkoreit, Jones, Gomez, Kaiser, and Polosukhin]{vaswani2017attention}
Ashish Vaswani, Noam Shazeer, Niki Parmar, Jakob Uszkoreit, Llion Jones, Aidan~N Gomez, {\L}ukasz Kaiser, and Illia Polosukhin.
\newblock Attention is all you need.
\newblock \emph{Advances in neural information processing systems}, 30, 2017.

\bibitem[Wang et~al.(2021{\natexlab{a}})Wang, Mallya, and Liu]{wang2021one}
Ting-Chun Wang, Arun Mallya, and Ming-Yu Liu.
\newblock One-shot free-view neural talking-head synthesis for video conferencing.
\newblock In \emph{Proceedings of the IEEE/CVF conference on computer vision and pattern recognition}, pages 10039--10049, 2021{\natexlab{a}}.

\bibitem[Wang et~al.(2021{\natexlab{b}})Wang, Yang, Bremond, and Dantcheva]{wang2021latent}
Yaohui Wang, Di Yang, Francois Bremond, and Antitza Dantcheva.
\newblock Latent image animator: Learning to animate images via latent space navigation.
\newblock In \emph{International Conference on Learning Representations}, 2021{\natexlab{b}}.

\bibitem[Wolf et~al.(2019)Wolf, Debut, Sanh, Chaumond, Delangue, Moi, Cistac, Rault, Louf, Funtowicz, et~al.]{wolf2019huggingface}
Thomas Wolf, Lysandre Debut, Victor Sanh, Julien Chaumond, Clement Delangue, Anthony Moi, Pierric Cistac, Tim Rault, R{\'e}mi Louf, Morgan Funtowicz, et~al.
\newblock Huggingface's transformers: State-of-the-art natural language processing.
\newblock \emph{arXiv preprint arXiv:1910.03771}, 2019.

\bibitem[Wood et~al.(2021)Wood, Baltru{\v{s}}aitis, Hewitt, Dziadzio, Cashman, and Shotton]{wood2021fake}
Erroll Wood, Tadas Baltru{\v{s}}aitis, Charlie Hewitt, Sebastian Dziadzio, Thomas~J Cashman, and Jamie Shotton.
\newblock Fake it till you make it: face analysis in the wild using synthetic data alone.
\newblock In \emph{Proceedings of the IEEE/CVF international conference on computer vision}, pages 3681--3691, 2021.

\bibitem[Wood et~al.(2022)Wood, Baltru{\v{s}}aitis, Hewitt, Johnson, Shen, Milosavljevi{\'c}, Wilde, Garbin, Sharp, Stojiljkovi{\'c}, et~al.]{wood20223d}
Erroll Wood, Tadas Baltru{\v{s}}aitis, Charlie Hewitt, Matthew Johnson, Jingjing Shen, Nikola Milosavljevi{\'c}, Daniel Wilde, Stephan Garbin, Toby Sharp, Ivan Stojiljkovi{\'c}, et~al.
\newblock 3d face reconstruction with dense landmarks.
\newblock In \emph{European Conference on Computer Vision}, pages 160--177. Springer, 2022.

\bibitem[Wu et~al.(2022)Wu, Liang, Ji, Yang, Fang, Jiang, and Duan]{wu2022nuwa}
Chenfei Wu, Jian Liang, Lei Ji, Fan Yang, Yuejian Fang, Daxin Jiang, and Nan Duan.
\newblock N{\"u}wa: Visual synthesis pre-training for neural visual world creation.
\newblock In \emph{European conference on computer vision}, pages 720--736. Springer, 2022.

\bibitem[Xie et~al.(2022)Xie, Zhang, Cao, Lin, Bao, Yao, Dai, and Hu]{xie2022simmim}
Zhenda Xie, Zheng Zhang, Yue Cao, Yutong Lin, Jianmin Bao, Zhuliang Yao, Qi Dai, and Han Hu.
\newblock Simmim: A simple framework for masked image modeling.
\newblock In \emph{Proceedings of the IEEE/CVF Conference on Computer Vision and Pattern Recognition}, pages 9653--9663, 2022.

\bibitem[Yan et~al.(2021)Yan, Zhang, Abbeel, and Srinivas]{yan2021videogpt}
Wilson Yan, Yunzhi Zhang, Pieter Abbeel, and Aravind Srinivas.
\newblock Videogpt: Video generation using vq-vae and transformers.
\newblock \emph{arXiv preprint arXiv:2104.10157}, 2021.

\bibitem[Zhu et~al.(2020)Zhu, Min, Kadav, and Graf]{zhu2020s3vae}
Yizhe Zhu, Martin~Renqiang Min, Asim Kadav, and Hans~Peter Graf.
\newblock S3vae: Self-supervised sequential vae for representation disentanglement and data generation.
\newblock In \emph{Proceedings of the IEEE/CVF Conference on Computer Vision and Pattern Recognition}, pages 6538--6547, 2020.

\end{thebibliography}
